\documentclass[11pt,a4paper]{article}
\usepackage{times,latexsym}
\usepackage{url}
\usepackage[T1]{fontenc}
\usepackage[utf8]{inputenc}

\usepackage{tikz}
\usepackage{subcaption}
\usepackage{colortbl}
\usepackage{makecell}
\usepackage{multicol}
\usepackage{float}
\usepackage{lscape}
\usepackage{multirow}
\usepackage{tabularx}
\usepackage{booktabs}
\usepackage{colortbl}
\usepackage{enumitem}
\usepackage{algorithm}
\usepackage{algorithmic}
\usepackage{longtable}
\usepackage{tabu}
\usepackage{longtable}
\usepackage{arydshln}
\usepackage{inconsolata}
\usepackage{rotating}

\usepackage{pdflscape}

\usepackage{soul}
\definecolor{HallRedShade}{HTML}{fdf1ec}
\definecolor{HallRedText}{HTML}{b76039}
\sethlcolor{HallRedShade}

\makeatletter
\def\adl@drawiv#1#2#3{%
        \hskip.5\tabcolsep
        \xleaders#3{#2.5\@tempdimb #1{1}#2.5\@tempdimb}%
                #2\z@ plus1fil minus1fil\relax
        \hskip.5\tabcolsep}
\newcommand{\cdashlinelr}[1]{%
  \noalign{\vskip 1.3pt
           \global\let\@dashdrawstore\adl@draw
           \global\let\adl@draw\adl@drawiv}
  \cdashline{#1}[.4pt/2pt]
  \noalign{\global\let\adl@draw\@dashdrawstore
           \vskip 1.3pt}}
\makeatother
\usepackage[acceptedWithA]{tacl2021v1}

\usepackage{xspace,mfirstuc,tabulary}

\newif\iftaclinstructions
\taclinstructionsfalse 
\iftaclinstructions
\renewcommand{\confidential}{}
\renewcommand{\anonsubtext}{(No author info supplied here, for consistency with
TACL-submission anonymization requirements)}
\newcommand{\instr}
\fi

\iftaclpubformat 

\else

\fi

\usepackage{soul}
\usepackage{pdflscape}
\definecolor{hl_color_da}{HTML}{FFC6C6}
\definecolor{hl_color_ng}{HTML}{D7C9FF}
\definecolor{hl_color_pc}{HTML}{FFD4AA}
\definecolor{hl_color_jw}{HTML}{BCE6E6}
\definecolor{hl_color_am}{HTML}{F7DCB4}
\definecolor{hl_color_bh}{HTML}{F7DCB4}
\definecolor{hl_color_ab}{HTML}{F7DCB4}
\definecolor{graytext}{HTML}{404040}

\definecolor{tikz_blue}{HTML}{009ADE}
\definecolor{tikz_red}{HTML}{F67280}
\definecolor{tikz_orange}{HTML}{F28522}
\definecolor{tikz_gray}{HTML}{E2E2E2}
\definecolor{tikz_green}{HTML}{91B54D}

\title{Hallucinations in Large Multilingual Translation Models}

\definecolor{citeblue}{HTML}{120d77}

\author{Nuno M. Guerreiro$^{1,3}$ \quad Duarte M. Alves$^{1,3}$ \quad Jonas Waldendorf\,$^{4}$ \\
\textbf{Barry Haddow}$^{4}$ \quad \textbf{Alexandra Birch}$^{4}$ \quad \textbf{Pierre Colombo}$^{5}$ \quad \textbf{André F. T. Martins}$^{1,2,3}$ \\
$^1$Instituto de Telecomunicações, Lisbon, Portugal \hfill $^2$Unbabel, Lisbon, Portugal\\
$^3$Instituto Superior T\'ecnico, University of Lisbon, Portugal\\
$^4$School of Informatics, University of Edinburgh\\
$^5$MICS, CentraleSupélec, Université Paris-Saclay\\
\small \url{miguelguerreironuno@gmail.com}}

\date{}

\begin{document}
\maketitle
\begin{abstract}
    Large-scale multilingual machine translation systems have demonstrated remarkable ability to translate directly between numerous languages, making them increasingly appealing for real-world applications. However, when deployed \textit{in the wild}, these models may generate hallucinated translations which have the potential to severely undermine user trust and raise safety concerns. Existing research on hallucinations has primarily focused on small bilingual models trained on high-resource languages, leaving a gap in our understanding of hallucinations in massively multilingual models across diverse translation scenarios. In this work, we fill this gap by conducting a comprehensive analysis on both the M2M family of conventional neural machine translation models and ChatGPT, a general-purpose large language model~(LLM) that can be prompted for translation. Our investigation covers a broad spectrum of conditions, spanning over 100 translation directions across various resource levels and going beyond English-centric language pairs. We provide key insights regarding the prevalence, properties, and mitigation of hallucinations, paving the way towards more responsible and reliable machine translation systems. 
\end{abstract}

\section{Introduction}
\label{sec:introduction}
Recent advancements in large-scale multilingual machine translation have brought us closer to realizing a universal translation system: a single model capable of handling numerous languages and translation directions~\citep{aharoni-etal-2019-massively, arivazhagan-etal-2019, fanetal2020_m2m, zhang-etal-2020-improving, wenzek-etal-2021-findings, goyal-etal-2022-flores, nllb2022}. Concurrently, general-purpose large language models~(LLMs) have exhibited a remarkable ability to generalize to new tasks, including translation, where they are becoming increasingly stronger~\citep{browngpt3_openai, palm_google, hendyetal2023_microsoftchatgpt}. Compared to traditional bilingual models, these systems can offer significant performance improvements and greatly simplify engineering efforts, as a single model can be used for all language pairs~\citep{arivazhagan-etal-2019}. As a result, they are an increasingly attractive choice for real-world applications. However, when deployed \textit{in the wild}, these models may still generate \textit{hallucinations}:~highly pathological translations that can severely damage user trust and pose serious safety concerns~\citep{perez2022red, kumaretal2022_harmeacl}.

The problem of hallucinations has long been recognized by researchers~\citep{jietal2022}, and recent studies have contributed towards better understanding, detection and mitigation of these pathological translations. However, these studies have been conducted on \textit{small bilingual} models~(<100M parameters) trained on a \textit{single English-centric high-resource} language pair~\citep{raunak-etal-2021-curious, ferrando-etal-2022-towards, guerreiro-etal-2022-lookingforaneedle, guerreiro-etal-2022-optimaltransport, dale-etal-2022-hallsalti, xu-etal-2022-taclhalls}. This leaves a knowledge gap regarding the prevalence and properties of hallucinations in large-scale translation models across different translation directions, domains and data conditions.


In this work, we aim to fill this gap by investigating hallucinations on two different classes of models. The first and main class in our analysis is the \textit{de facto} standard approach of massively multilingual supervised models: we use the M2M-100 family of multilingual NMT models~\citep{fanetal2020_m2m}, which includes the largest open-source multilingual NMT model with 12B parameters. The second class is the novel and promising approach of leveraging generative LLMs for translation. Contrary to conventional NMT models, these models are trained on massive amounts of monolingual data in many languages, with a strong bias towards English, and do not require parallel data. In our analysis, we use ChatGPT\footnote{\url{https://openai.com/blog/chatgpt}; this system has not been documented, so details of training data and training regime are unknown.}, a LLM that has been shown to achieve surprisingly high translation quality over a wide range of language pairs~\citep{hendyetal2023_microsoftchatgpt, Peng2023ChatGPT4MT}.

We organize our study by analyzing the two prevalent types of hallucinations in NMT considered in the literature: hallucinations under perturbation and natural hallucinations~\citep{Lee2018HallucinationsIN, raunak-etal-2021-curious, guerreiro-etal-2022-lookingforaneedle}. Firstly, we study hallucinations under perturbation and evaluate whether these translation systems are robust to source-side artificial perturbations. While previous studies have found that these perturbations (e.g., spelling errors and capitalization mistakes) can reliably induce hallucinations~\citep{Lee2018HallucinationsIN, raunak-etal-2021-curious}, it is not clear whether those conclusions hold for large multilingual models. Secondly, we comprehensively investigate natural hallucinations, and evaluate their prevalence and properties in the outputs of the massively multilingual M2M models on a vast range of conditions, spanning from English-centric to non-English-centric language pairs, translation directions with little supervision, and specialized sensitive domains where hallucinations have devastating impact on user trust (e.g., medical data). Finally, we study a hybrid setup where other translation systems can be requested as fallback systems when an original system hallucinates, with the aim of mitigating hallucinations and improving overall translation quality. 

Our analysis reveals several key insights on the prevalence and properties of hallucinations, including:
\begin{itemize}
    \item multilingual models predominantly struggle with hallucinations in low-resource language pairs and translating out of English, with hallucination rates well above 10\% for some translation directions;
    \item hallucinations in low-resource language pairs can manifest toxic patterns that can be traced back to the training data, posing serious safety issues;
    \item smaller distilled models can mitigate hallucinations by incorporating modeling choices that discourage them, such as leveraging less potent shallow decoders that rely more on the encoder representations, and reducing bias towards higher-resource language pairs through uniform sampling of translation directions during distillation;
    \item ChatGPT produces hallucinations that are qualitatively different from those of conventional translation models, mostly consisting of off-target translations, overgeneration, and even failed attempts to translate;
    \item hallucinations are \textit{sticky} and hard to reverse with models that share the same training data and architecture, whereas employing more diverse models as fallback systems can substantially improve overall translation quality and eliminate pathologies like oscillatory hallucinations.
\end{itemize}

We release all our code and make available over a million translations in more than 100 translation directions to spur future research.\footnote{All resources will be made available in \url{https://github.com/deep-spin/lmt_hallucinations}.}

\section{Background}
\label{sec:background}
\subsection{Large Multilingual Language Models}
Massively multilingual neural machine translation has recently emerged as a powerful paradigm for building machine translation systems that can handle numerous languages~\citep{akhbardeh-etal-2021-findings,wenzek-etal-2021-findings,nllb2022, siddhantetal2022_multilingual_google, googletranslate2022, palm_google}. These systems aim to translate directly with a single model for multiple language pairs without relying on any pivot language.

The dominant strategy for achieving these systems is to train large multilingual models on vast amounts of parallel data often obtained through a combination of data mining and data augmentation strategies, such as backtranslation~\citep{sennrich-etal-2016-improving, edunov-etal-2018-understanding}. Compared to classic bilingual models, the multilinguality of these systems results in significant improvements, particularly for low-resource and non-English-centric language pairs, as these benefit the most from multilingual transfer~\citep{arivazhagan-etal-2019, fanetal2020_m2m}. 

As an alternative, a novel and promising strategy is to leverage the emergent capabilities of large language models (LLMs). These systems are pretrained on massive nonparallel corpora and can be prompted to solve arbitrary tasks~\citep{radford2019language, browngpt3_openai}. In fact, this approach has led to impressive results across a wide variety of NLP tasks~\citep{palm_google, opt_2022}. Translation is no exception: LLMs can produce fluent and adequate translations, especially for high-resource English-centric language pairs, that are competitive with those of dedicated supervised translation models~\citep{vilaretal2023_palm, Peng2023ChatGPT4MT, garciaetal2023_google, hendyetal2023_microsoftchatgpt, bawdenandyvon2023_bloom}.

\subsection{Hallucinations in Machine Translation}
Hallucinations lie at the extreme end of translation pathologies and present a critical challenge in machine translation, as they can severely compromise the safety and reliability of real-world applications.

Importantly, hallucinations in machine translation are unlike hallucinations in other natural language generation tasks (e.g., abstractive summarization and generative question answering)~\citep{jietal2022}. While, for these other tasks, models often produce hallucinated outputs~\citep{falke-etal-2019-ranking, cao-etal-2022-hallucinated, selfcheckgpt_2023}, hallucinations in machine translation, possibly attributed to the more closed-ended nature of the task, are substantially rarer and hard to observe in clean, unperturbed data. This has led several previous studies to examine their properties by creating artificial scenarios where hallucinations are more likely to occur (e.g., introducing perturbations in the source text~\citep{Lee2018HallucinationsIN} or noise in the training data~\citep{raunak-etal-2021-curious}). To distinguish these two scenarios, hallucinations in machine translation are categorized into two types~\citep{raunak-etal-2021-curious}: \textit{hallucinations under perturbation} and \textit{natural hallucinations}.

\paragraph{Hallucinations under perturbation.} A model generates a hallucination under perturbation when it produces a significantly lower quality translation for a slightly perturbed input compared to the original input~\citep{Lee2018HallucinationsIN}. Hallucinations under perturbation explicitly reveal the lack of robustness of translation systems to perturbations in the source text~(e.g., misspellings or capitalization errors) by finding translations that undergo significant negative shifts in quality due to these changes.

\paragraph{Natural hallucinations.} Contrary to hallucinations under perturbations, these translations occur naturally without any perturbation. As a result, natural hallucinations are rare and challenging to study. In this work, we follow the taxonomy introduced in~\citet{raunak-etal-2021-curious} and later extended in~\citet{guerreiro-etal-2022-lookingforaneedle}. Under this taxonomy, hallucinations are translations that contain content that is detached from the source text. To distinguish between different types of hallucinations, they can be categorized as \textit{largely fluent detached hallucinations} or \textit{oscillatory hallucinations}. The former refers to translations that bear minimal or no relation at all to the source, while the latter refers to inadequate translations that contain erroneous repetitions of words and phrases.

\section{Experimental Suite}
\label{sec:experimental_suite}
In this section, we provide an overview of the models, datasets and evaluation metrics used throughout our study. 
\subsection{Models}
We focus on two classes of models: (i) conventional supervised multilingual NMT models, and (ii) LLMs that can be prompted for translation.

For the supervised multilingual NMT models, we use the transformer-based~\citep{transformer_vaswani} M2M-100 family of models~\citep{fanetal2020_m2m}, which consists of three variants with different sizes: \texttt{M2M~(S)} with 418M parameters, \texttt{M2M~(M)} with 1.2B parameters, and \texttt{M2M~(L)}~--- the largest available open-source multilingual NMT model~---  with 12B parameters. These models were trained on a many-to-many parallel dataset comprising 7.5B sentences crawled from the web, and support 100 languages and thousands of translation directions. We also experiment with \texttt{SMaLL100} \citep{mohammadshahi-etal-2022-small}, a shallow multilingual NMT model with 330M parameters obtained via distillation of \texttt{M2M (L)}. Unlike the M2M models, \texttt{SMaLL100} was trained on a much smaller training set with uniform sampling across all language pairs to reduce the bias towards high-resource languages: only 100k parallel sentences from the original M2M training data were used for each translation direction, for a total of 456M parallel sentences. For decoding, we run beam search with a beam size of 4. All experiments were run on \texttt{fairseq}~\citep{ott-etal-2019-fairseq}. 

As for the alternative strategy using LLMs, we use \texttt{ChatGPT} (\texttt{gpt-3.5-turbo})\footnote{\url{https://platform.openai.com/docs/models/gpt-3-5}; we used the API in March, 2023.} , a variant of GPT3.5 --- a GPT-family~\citep{Radford2018ImprovingLU, radford2019language, browngpt3_openai} large-scale model with 175B parameters --- that has been fine-tuned with human feedback in the style of InstructGPT~\citep{instructgpt_2022}. \texttt{ChatGPT} has been shown to achieve impressive results for multiple multilingual NLP tasks, including translation~\citep{kocmiandfedermann2023, lu2023error, gptscore_2023, hendyetal2023_microsoftchatgpt, Peng2023ChatGPT4MT}. To generate translations, we use the zero-shot prompt template used in ~\citet{hendyetal2023_microsoftchatgpt} and keep the generation parameters as the default API parameters.\footnote{We encountered several API/server errors when prompting \texttt{ChatGPT} for translation with temperature 0, particularly for low-resource language pairs and languages with lower coverage scripts. Those errors are alleviated, although not entirely eliminated, when the default parameters are used.}

\subsection{Datasets}
\label{subsec:datasets}
We carefully selected datasets based on two main criteria: their familiarity to researchers and practitioners, and the avoidance of train/test overlap for the M2M models.\footnote{\texttt{ChatGPT}’s training data is not publicly available. As such, we cannot guarantee that it has not been exposed to the data we use in our analysis.} To this end, we chose to use premier translation benchmarks: \textsc{Flores}-101~\citep{goyal-etal-2022-flores}, WMT and TICO~\citep{anastasopoulos-etal-2020-tico}. \textsc{Flores}-101 is a high-quality multi-parallel dataset that consists of Wikipedia text in 101 languages and allows for the assessment of hallucinations across a vast range of translation directions; we join the \texttt{dev} and \texttt{devtest} subsets for evaluation. For WMT, we used the same benchmarks as those used in the original M2M paper evaluation suite, as these were explicitly removed from the training data. Additionally, we selected recent WMT test sets from the WMT21 and WMT22 campaigns as they were released after the models were trained. In contrast to these general-purpose datasets, TICO is a specialized medical-domain multilingual benchmark that includes COVID-19 related data, such as medical papers and news articles; we join the \texttt{dev} and \texttt{test} sets. Full details about the datasets can be found in Appendix~\ref{app:datasets}.

\subsection{Evaluation Metrics}
Throughout our work, we focus mainly on sentence-level evaluation. Our main lexical metric is spBLEU~\citep{goyal-etal-2022-flores},\footnote{We use spBLEU as implemented in \textsc{Sacrebleu}~\citep{post-2018-call}: \texttt{nrefs:1|case:mixed|eff:yes|tok:flores101|\\smooth:exp|version:2.3.1}.} as it has been widely employed in works on massively multilingual translation~\citep{fanetal2020_m2m, wenzek-etal-2021-findings, mohammadshahi-etal-2022-small, nllb2022} and offers fairer evaluation for low-resource languages compared to BLEU~\citep{papineni-etal-2002-bleu}. Moreover, we follow the most recent MT metrics shared-task recommendations~\citep{freitag-etal-2022-results} and also adopt neural metrics. We use the latest reference-based and reference-free COMET variants: COMET-22~\citep{rei-etal-2022-comet} and CometKiwi~\citep{rei-etal-2022-cometkiwi}. Lastly, we use the cross-lingual encoder LaBSE~\citep{feng-etal-2022-language} to obtain sentence similarity scores, as these have been successfully employed in prior research on detection of natural hallucinations~\citep{guerreiro-etal-2022-optimaltransport, dale-etal-2022-hallsalti}.

\section{Hallucinations under Perturbation}
\label{sec:halls_under_perturbation}
We start our analysis by focusing on artificially created hallucinations. We first provide an overview of our experimental setting, focusing on the construction of the perturbed data and detection approach. Then, we present our results and analyze the properties of these hallucinations across different resource levels and models.

\begin{table*}[t]
\centering
\renewcommand\arraystretch{0.9}
\footnotesize
\begin{tabular}{>{\arraybackslash}m{1.45cm} >{\arraybackslash}m{1.5cm} >{\arraybackslash}m{1.15cm} c >{\raggedleft\arraybackslash}m{1.75cm} >{\arraybackslash}m{1.15cm} c >{\arraybackslash}m{1.5cm} >{\arraybackslash}m{1.15cm}}
\toprule
\multirow{2}{*}{\textsc{Model}} & \multicolumn{2}{c}{\textbf{\textsc{Low Resource}}} & & \multicolumn{2}{c}{\textbf{\textsc{Mid Resource}}} & & \multicolumn{2}{c}{\textbf{\textsc{High Resource}}}\\\cmidrule{2-3}\cmidrule{5-6}\cmidrule{8-9}
& LP Fraction & Rate (\%) & & \multicolumn{1}{l}{{LP Fraction}} & Rate (\%) & & LP Fraction & Rate (\%)\\\midrule
\texttt{SMaLL100} & \begin{tikzpicture}[font=\scriptsize] 
 \draw [fill=tikz_red] (0,0) rectangle (2*6/7*0.15, 0.15); 
 \draw [fill=tikz_gray] (2*6/7*0.15, 0) rectangle (6*0.15, 0.15); 
 \node [left, color=tikz_red] at (0, 0.1) {\textbf{\scriptsize{2}}\textcolor{black!50}{/7}}; 
 \end{tikzpicture} & $0.213_{\textcolor{black!50}{\tiny{0.00}}}$ & &\begin{tikzpicture}[font=\scriptsize] 
 \draw [fill=tikz_red] (0,0) rectangle (2*6/19*0.15, 0.15); 
 \draw [fill=tikz_gray] (2*6/19*0.15, 0) rectangle (6*0.15, 0.15); 
 \node [left, color=tikz_red] at (0, 0.1) {\textbf{\scriptsize{2}}\textcolor{black!50}{/19}}; 
 \end{tikzpicture} & $0.009_{\textcolor{black!50}{\tiny{0.00}}}$ & &\begin{tikzpicture}[font=\scriptsize] 
 \draw [fill=tikz_red] (0,0) rectangle (1*6/5*0.15, 0.15); 
 \draw [fill=tikz_gray] (1*6/5*0.15, 0) rectangle (6*0.15, 0.15); 
 \node [left, color=tikz_red] at (0, 0.1) {\textbf{\scriptsize{1}}\textcolor{black!50}{/5}}; 
 \end{tikzpicture} & $0.017_{\textcolor{black!50}{\tiny{0.00}}}$ \\ \texttt{M2M (S)} & \begin{tikzpicture}[font=\scriptsize] 
 \draw [fill=tikz_red] (0,0) rectangle (5*6/7*0.15, 0.15); 
 \draw [fill=tikz_gray] (5*6/7*0.15, 0) rectangle (6*0.15, 0.15); 
 \node [left, color=tikz_red] at (0, 0.1) {\textbf{\scriptsize{5}}\textcolor{black!50}{/7}}; 
 \end{tikzpicture} & $0.261_{\textcolor{black!50}{\tiny{0.08}}}$ & &\begin{tikzpicture}[font=\scriptsize] 
 \draw [fill=tikz_red] (0,0) rectangle (11*6/19*0.15, 0.15); 
 \draw [fill=tikz_gray] (11*6/19*0.15, 0) rectangle (6*0.15, 0.15); 
 \node [left, color=tikz_red] at (0, 0.1) {\textbf{\scriptsize{11}}\textcolor{black!50}{/19}}; 
 \end{tikzpicture} & $0.140_{\textcolor{black!50}{\tiny{0.08}}}$ & &\begin{tikzpicture}[font=\scriptsize] 
 \draw [fill=tikz_red] (0,0) rectangle (0*6/5*0.15, 0.15); 
 \draw [fill=tikz_gray] (0*6/5*0.15, 0) rectangle (6*0.15, 0.15); 
 \node [left, color=tikz_red] at (0, 0.1) {\textbf{\scriptsize{0}}\textcolor{black!50}{/5}}; 
 \end{tikzpicture} & $0.000_{\textcolor{black!50}{\tiny{0.00}}}$ \\ \texttt{M2M (M)} & \begin{tikzpicture}[font=\scriptsize] 
 \draw [fill=tikz_red] (0,0) rectangle (3*6/7*0.15, 0.15); 
 \draw [fill=tikz_gray] (3*6/7*0.15, 0) rectangle (6*0.15, 0.15); 
 \node [left, color=tikz_red] at (0, 0.1) {\textbf{\scriptsize{3}}\textcolor{black!50}{/7}}; 
 \end{tikzpicture} & $0.083_{\textcolor{black!50}{\tiny{0.00}}}$ & &\begin{tikzpicture}[font=\scriptsize] 
 \draw [fill=tikz_red] (0,0) rectangle (6*6/19*0.15, 0.15); 
 \draw [fill=tikz_gray] (6*6/19*0.15, 0) rectangle (6*0.15, 0.15); 
 \node [left, color=tikz_red] at (0, 0.1) {\textbf{\scriptsize{6}}\textcolor{black!50}{/19}}; 
 \end{tikzpicture} & $0.035_{\textcolor{black!50}{\tiny{0.00}}}$ & &\begin{tikzpicture}[font=\scriptsize] 
 \draw [fill=tikz_red] (0,0) rectangle (0*6/5*0.15, 0.15); 
 \draw [fill=tikz_gray] (0*6/5*0.15, 0) rectangle (6*0.15, 0.15); 
 \node [left, color=tikz_red] at (0, 0.1) {\textbf{\scriptsize{0}}\textcolor{black!50}{/5}}; 
 \end{tikzpicture} & $0.000_{\textcolor{black!50}{\tiny{0.00}}}$ \\ \texttt{M2M (L)} & \begin{tikzpicture}[font=\scriptsize] 
 \draw [fill=tikz_red] (0,0) rectangle (4*6/7*0.15, 0.15); 
 \draw [fill=tikz_gray] (4*6/7*0.15, 0) rectangle (6*0.15, 0.15); 
 \node [left, color=tikz_red] at (0, 0.1) {\textbf{\scriptsize{4}}\textcolor{black!50}{/7}}; 
 \end{tikzpicture} & $0.296_{\textcolor{black!50}{\tiny{0.08}}}$ & &\begin{tikzpicture}[font=\scriptsize] 
 \draw [fill=tikz_red] (0,0) rectangle (3*6/19*0.15, 0.15); 
 \draw [fill=tikz_gray] (3*6/19*0.15, 0) rectangle (6*0.15, 0.15); 
 \node [left, color=tikz_red] at (0, 0.1) {\textbf{\scriptsize{3}}\textcolor{black!50}{/19}}; 
 \end{tikzpicture} & $0.017_{\textcolor{black!50}{\tiny{0.00}}}$ & &\begin{tikzpicture}[font=\scriptsize] 
 \draw [fill=tikz_red] (0,0) rectangle (0*6/5*0.15, 0.15); 
 \draw [fill=tikz_gray] (0*6/5*0.15, 0) rectangle (6*0.15, 0.15); 
 \node [left, color=tikz_red] at (0, 0.1) {\textbf{\scriptsize{0}}\textcolor{black!50}{/5}}; 
 \end{tikzpicture} & $0.000_{\textcolor{black!50}{\tiny{0.00}}}$ \\\cdashlinelr{1-9} \texttt{ChatGPT} & \begin{tikzpicture}[font=\scriptsize] 
 \draw [fill=tikz_red] (0,0) rectangle (4*6/7*0.15, 0.15); 
 \draw [fill=tikz_gray] (4*6/7*0.15, 0) rectangle (6*0.15, 0.15); 
 \node [left, color=tikz_red] at (0, 0.1) {\textbf{\scriptsize{4}}\textcolor{black!50}{/7}}; 
 \end{tikzpicture} & $0.059_{\textcolor{black!50}{\tiny{0.08}}}$ & &\begin{tikzpicture}[font=\scriptsize] 
 \draw [fill=tikz_red] (0,0) rectangle (10*6/19*0.15, 0.15); 
 \draw [fill=tikz_gray] (10*6/19*0.15, 0) rectangle (6*0.15, 0.15); 
 \node [left, color=tikz_red] at (0, 0.1) {\textbf{\scriptsize{10}}\textcolor{black!50}{/19}}; 
 \end{tikzpicture} & $0.183_{\textcolor{black!50}{\tiny{0.08}}}$ & &\begin{tikzpicture}[font=\scriptsize] 
 \draw [fill=tikz_red] (0,0) rectangle (0*6/5*0.15, 0.15); 
 \draw [fill=tikz_gray] (0*6/5*0.15, 0) rectangle (6*0.15, 0.15); 
 \node [left, color=tikz_red] at (0, 0.1) {\textbf{\scriptsize{0}}\textcolor{black!50}{/5}}; 
 \end{tikzpicture} & $0.000_{\textcolor{black!50}{\tiny{0.00}}}$ \\ \bottomrule
\end{tabular}
\caption{Fraction of languages for which models produces at least one hallucination under perturbation, and average hallucination rate (and median, in subscript) across all languages at each resource level.}
\label{tab:flores_halls_under_perturb_20}
\end{table*}

\begin{figure*}
    \centering
    \includegraphics[width=\linewidth]{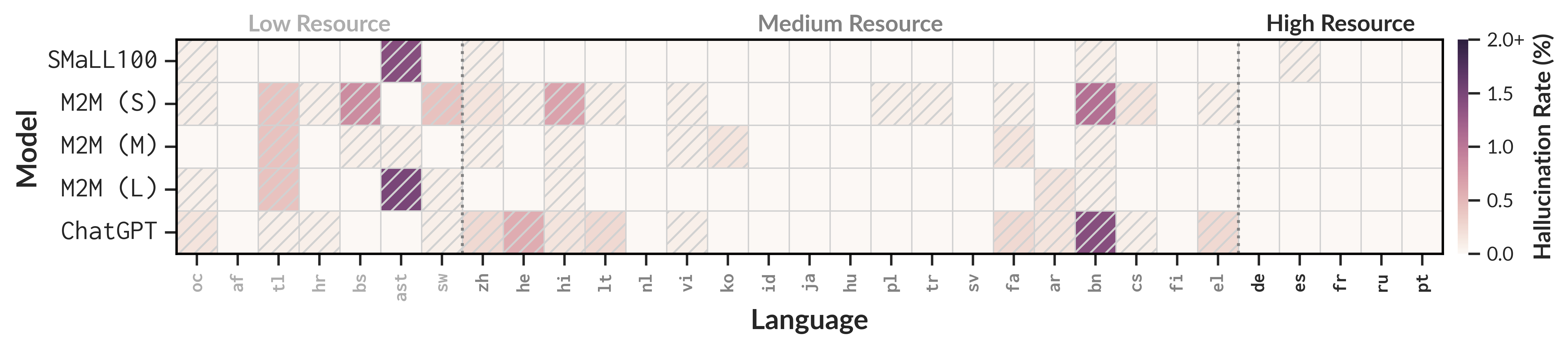}\vspace{-8pt} \caption{Heatmap of hallucination rates for each model in the languages considered. Pattern-filled cells indicate at least one hallucination under perturbation for a given model-language pair.}
    \label{fig:flores_halls_under_perturb_20}
\end{figure*}

\subsection{Evaluation Setting}
\label{subsec:halls_under_perturb_eval_setting}

\paragraph{Perturbations.} To construct the perturbed source sequences, we apply the same minimal perturbations used in \citet{xu-etal-2022-taclhalls}: misspeling of words, insertion of frequent tokens in the beginning of the source sequence, and capitalization errors. For full details on the construction of the perturbed data, refer to Appendix~\ref{app:data_for_halls_under_perturbation}.

\paragraph{Translation directions.} We use the \textsc{Flores} dataset for these experiments, and focus specifically on translation out of English. We selected all bridge languages\footnote{In the M2M paper~\citep{fanetal2020_m2m}, a bridge language is defined as a language that connects languages across language groupings (e.g., Romance, Slavic languages) and is mined against all other bridge languages. It is usually one of the most resourced languages within a language grouping and its purpose is to reduce the number of bitext pairs while preserving translation directions of practical interest.}, as well as additional low-resource languages that were underrepresented among bridge languages. Overall, we generate translations for 31 different language pairs~(LPs). We present the language pairs and more details on our choice of languages in Appendix~\ref{app:data_for_halls_under_perturbation}.

\paragraph{Detection.} Our detection approach is inspired by that of previous works on hallucinations under perturbation~\citep{Lee2018HallucinationsIN, raunak-etal-2021-curious, ferrando-etal-2022-towards, xu-etal-2022-taclhalls}. The algorithm is a simple 2-rule process: we fix (i)~a minimum threshold quality score for the original translations, and (ii)~an extremely low maximum quality score for the perturbed translations. A model generates a hallucination under perturbation when both translations meet the thresholds. Crucially, rule (i) ensures that low-quality translations for unperturbed sources are not considered as candidates for hallucinations under perturbation.\footnote{Note that low-quality translations for unperturbed sources fall under the scope of the study on natural hallucinations, that follows in subsequent sections of the paper.}

We extend this algorithm to handle multiple models and language pairs by adapting rule (i). We first obtain source sentences for which all models produce translations that meet a minimum quality threshold (spBLEU > 9). Then, we sort them according to average quality across the different models, and select the top 20\% as candidates. Finally, we apply rule (ii) and set the threshold to spBLEU < 3. We selected both thresholds based on the choices made in previous works~\citep{raunak-etal-2021-curious, ferrando-etal-2022-towards, xu-etal-2022-taclhalls}.

This approach ensures a fixed sample size across different language pairs, and that the sentences analyzed for each language pair are consistent across all models. Moreover, it allows us to effectively detect hallucinations under perturbation across multiple models in a multilingual scenario in a scalable manner, while accounting for the unique quality trends observed across different models and languages.\footnote{Note that detection of hallucinations under perturbation does not explicitly target detachment from the source text. We provide a broader discussion on the difference between this detection approach and that of natural hallucinations~(introduced later in Section~\ref{subsec:natural_halls_evalsetting}) in Appendix~\ref{app:supp_results_for_hall_under_perturbation}.}

\subsection{Results}

We show aggregated results in Table~\ref{tab:flores_halls_under_perturb_20} and language-specific results in Figure~\ref{fig:flores_halls_under_perturb_20}. Overall, they reveal that perturbations have the potential to trigger hallucinations under perturbation, even in larger models. In what follows, we highlight several noteworthy trends found in our results.

\paragraph{Average hallucination rates generally decrease with increasing resource levels.} Table~\ref{tab:flores_halls_under_perturb_20} shows that all models, with the exception of \texttt{ChatGPT} that we analyze separately below, exhibit lower hallucination rates as resource levels increase. This is expected and suggests that models are better equipped to handle source-side perturbations for language pairs with more parallel data during training. In fact, hallucinations under perturbation for high-resource languages are almost non-existent. However, Figure~\ref{fig:flores_halls_under_perturb_20} reveals variability across languages, and even within the models in the same family that have been trained on the same data. For instance, when translating to Asturian~(\texttt{ast}), \texttt{M2M~(L)} and its distilled version \texttt{SMaLL100} have significantly higher hallucination rates than the smaller \texttt{M2M~(S)}. Thus, hallucinations under perturbation may emerge in other non-trivial ways unrelated to the training data.

\paragraph{\texttt{SMaLL100} exhibits lower hallucination rates than its teacher model \texttt{M2M (L)}.}  Recall that \texttt{SMaLL100} was trained using uniform sampling across all language pairs to prevent bias towards higher resourced language pairs. The results in Table~\ref{tab:flores_halls_under_perturb_20} may reflect one positive outcome from such approach: despite being much smaller than~\texttt{M2M~(L)}, \texttt{SMaLL100} hallucinates less and for fewer languages than its teacher model for low- and mid-resource language pairs.

\paragraph{Hallucinations under perturbation are not correlated with the quality of original translations.} The common approach for detection of hallucinations under perturbation~(see Section~\ref{subsec:halls_under_perturb_eval_setting}) raises an interesting question: \emph{are the original source sentences for which models produce higher quality translations less likely to lead to hallucinations when perturbed?} Our analysis found a very weak correlation~(according to Pearson correlation; see Appendix~\ref{app:supp_results_for_hall_under_perturbation}) between hallucinations under perturbation and spBLEU scores for the original unperturbed sources across all models. This indicates that even minimal perturbations in the source text can cause models to undergo significant shifts in translation quality.

\paragraph{\texttt{ChatGPT} exhibits different hallucination patterns from conventional translation models.} Table~\ref{tab:flores_halls_under_perturb_20} shows that, contrary to traditional models, \texttt{ChatGPT} generates more hallucinations for mid-resource languages than for low-resource languages. In fact, it surprisingly produces fewer hallucinations for low-resource languages than any other model. Moreover, \texttt{ChatGPT}'s hallucinations are qualitatively different from those of other models: they often consist of off-target translations,\footnote{We perform automatic language identification using the \texttt{fasttext}~\citep{joulin2016fasttext} LID model \texttt{lid.176.bin}.} overgeneration, or even failed attempts to translate (e.g., \textit{“This is an English sentence, so there is no way to translate it to Vietnamese”}; we provide further examples in Appendix~\ref{app:supp_results_for_hall_under_perturbation}). Furthermore, unlike traditional NMT models that frequently produce oscillatory character hallucinations, \texttt{ChatGPT} does not generate any such hallucinations under perturbation. This is further evidence that translation errors, even severely critical ones, obtained via prompting a LLM are different from those produced by traditional machine translation models~\citep{vilaretal2023_palm, garciaetal2023_google, hendyetal2023_microsoftchatgpt, bawdenandyvon2023_bloom}.

Interestingly, we also found that the vast majority of the hallucinations can be reversed with further sampling from the model.\footnote{We also found this to be the case with a one-shot prompt.} This connects to findings in~\citet{guerreiro-etal-2022-lookingforaneedle, selfcheckgpt_2023}: as with traditional NMT models, hallucinations with a LLM may not necessarily indicate model defect or incapacity to generate adequate translations, and may just result from “bad luck” during generation.

\section{Natural Hallucinations}
\label{sec:natural_halls}

Let us now turn to investigating natural hallucinations.\footnote{From now on, we will use the terms natural hallucinations~--- both detached and oscillatory hallucinations --- and hallucinations interchangeably.} We first provide an in-depth overview of our evaluation setting, focusing on the scenarios and detection methodology. Subsequently, we present a thorough analysis, exploring diverse properties of natural hallucinations such as their different types, the influence of translation direction, and prevalence of toxicity.

\subsection{Evaluation Setting}
\label{subsec:natural_halls_evalsetting}
\paragraph{Evaluation scenarios.} Analyzing massively multilingual translation models opens up several research scenarios that have not been studied in previous works that focused solely on bilingual models. We will take advantage of this opportunity and investigate natural hallucinations in three different evaluation scenarios, studying more than 100 translation directions in the main text alone. 

We start with an English-centric scenario where we pair 32 different languages with English for a total of 64 translation directions. Then, we study a non-English-centric scenario inspired by~\citet{fanetal2020_m2m}, where we explore 25 language pairs corresponding to real-world use cases of translation not involving English (e.g., translating Greek directly to Turkish). Finally, we assess the prevalence of hallucinations on sensitive medical data where they can have a devastating impact on user trust. We pair 9 different languages with English for a total of 18 directions. We present all the translation directions investigated in these setups in Appendix~\ref{app:subsec_translation_directions}. We report results for the first two setups using the \textsc{Flores} dataset in the main text and WMT in Appendix~\ref{app:subsec:engcentric}. For the final setup, we use the medical-domain TICO dataset.

\paragraph{Detection.} We integrate key findings from recent research on detection of hallucinations and focus on two main detectors: ALTI+~\citep{ferrando-etal-2022-towards} for detached hallucinations, and top $n$-gram (TNG)~\citep{raunak-etal-2021-curious, salted_raunak2022, guerreiro-etal-2022-lookingforaneedle} for oscillatory hallucinations. 

ALTI+ evaluates the relative contributions of both source and target prefixes to model predictions. As hallucinations are translations detached from the source sequence, ALTI+ can effectively detect them by identifying sentences with minimal source contribution. Notably, it faithfully reflects model behavior and explicitly signals model detachment from the source text in any translation direction~\citep{ferrando-etal-2022-towards}. In previous works, this method has been successfully employed to detect hallucinated toxicity in a multilingual context in~\citet{nllb2022}, and it has been validated on human-annotated hallucinations in~\citet{dale-etal-2022-hallsalti}, where it was demonstrated that ALTI+ scores easily separate detached hallucinations from other translations.\footnote{We followed the recommendations in~\citet{guerreiro-etal-2022-lookingforaneedle} and set model-based ALTI+ thresholds based on validation data where the models are expected to perform well. Specifically, we obtained the lowest 0.02\% --- in line with natural hallucination rates reported in the literature~\citep{salted_raunak2022} --- of the ALTI+ score distributions for high-resource WMT benchmarks. Additionally, to ensure further trustworthy, high-precision measurements, we excluded detected candidates with LaBSE or CometKiwi scores --- as these have been also been validated for detection of human-annotated detached hallucinations~\citep{dale-etal-2022-hallsalti, guerreiro-etal-2022-optimaltransport} --- exceeding the top 10\% of scores on translations from the same WMT benchmarks.}

TNG, on the other hand, is a straightforward, lightweight black-box heuristic targeting oscillatory hallucinations. It works by comparing the count of the top repeated translation $n$-gram to the count of the top repeated source $n$-gram, ensuring the difference is at least $t$. This approach has been validated on human-annotated hallucinations and found to identify oscillatory hallucinations with perfect precision~\citep{guerreiro-etal-2022-lookingforaneedle}. We follow previous work by using $n=4$ and $t=2$~\citep{raunak-etal-2021-curious, guerreiro-etal-2022-lookingforaneedle} and excluding translations that meet the reasonable quality threshold outlined in Section~\ref{subsec:halls_under_perturb_eval_setting}.\footnote{Note that oscillatory hallucinations can be simultaneously detected with ALTI+ and TNG.}

\paragraph{Remark on Model Selection.} We rely on ALTI+, a model-based detector, for reliable detection of detached hallucinations. Since we lack access to glass-box internal features from \texttt{ChatGPT}, we exclude it from our model selection to ensure consistency in our analysis. It is important to note that using alternative detectors could lead to misleading results and create discrepancies between the evaluation scenarios for \texttt{ChatGPT} and other models. Nonetheless, we will further examine \texttt{ChatGPT} in Section~\ref{sec:mitigation}, exploring various aspects such as the generation of oscillatory hallucinations and translation quality in scenarios where other models produce hallucinations.

\begin{table*}[t]
\centering
\renewcommand\arraystretch{0.9}
\footnotesize
\begin{tabular}{>{\arraybackslash}m{1.45cm} >{\arraybackslash}m{1.75cm} >{\arraybackslash}m{1.15cm} c >{\raggedright\arraybackslash}m{1.75cm} >{\arraybackslash}m{1.15cm} c >{\arraybackslash}m{1.85cm} >{\arraybackslash}m{1.15cm}}
\toprule
\multirow{2}{*}{\textsc{Model}} & \multicolumn{2}{c}{\textbf{\textsc{Low Resource}}} & & \multicolumn{2}{c}{\textbf{\textsc{Mid Resource}}} & & \multicolumn{2}{c}{\textbf{\textsc{High Resource}}}\\\cmidrule{2-3}\cmidrule{5-6}\cmidrule{8-9}
& LP Fraction & Rate (\%) & & \multicolumn{1}{l}{{LP Fraction}} & Rate (\%) & & LP Fraction & Rate (\%)\\\midrule
\texttt{SMaLL100} & \begin{tikzpicture}[font=\scriptsize] 
 \draw [fill=tikz_red] (0,0) rectangle (14*6/16*0.15, 0.15); 
 \draw [fill=tikz_gray] (14*6/16*0.15, 0) rectangle (6*0.15, 0.15); 
 \node [left, color=tikz_red] at (0, 0.1) {\textbf{\scriptsize{14}}\textcolor{black!50}{/16}}; 
 \end{tikzpicture} & $2.352_{\,\textcolor{black!50}{\tiny{0.57}}}$ & &\begin{tikzpicture}[font=\scriptsize] 
 \draw [fill=tikz_red] (0,0) rectangle (19*6/38*0.15, 0.15); 
 \draw [fill=tikz_gray] (19*6/38*0.15, 0) rectangle (6*0.15, 0.15); 
 \node [left, color=tikz_red] at (0, 0.1) {\textbf{\scriptsize{19}}\textcolor{black!50}{/38}}; 
 \end{tikzpicture} & $0.055_{\,\textcolor{black!50}{\tiny{0.02}}}$ & &\begin{tikzpicture}[font=\scriptsize] 
 \draw [fill=tikz_red] (0,0) rectangle (1*6/10*0.15, 0.15); 
 \draw [fill=tikz_gray] (1*6/10*0.15, 0) rectangle (6*0.15, 0.15); 
 \node [left, color=tikz_red] at (0, 0.1) {\textbf{\scriptsize{1}}\textcolor{black!50}{/10}}; 
 \end{tikzpicture} & $0.005_{\,\textcolor{black!50}{\tiny{0.00}}}$ \\ \texttt{M2M (S)} & \begin{tikzpicture}[font=\scriptsize] 
 \draw [fill=tikz_red] (0,0) rectangle (15*6/16*0.15, 0.15); 
 \draw [fill=tikz_gray] (15*6/16*0.15, 0) rectangle (6*0.15, 0.15); 
 \node [left, color=tikz_red] at (0, 0.1) {\textbf{\scriptsize{15}}\textcolor{black!50}{/16}}; 
 \end{tikzpicture} & $15.20_{\,\textcolor{black!50}{\tiny{2.86}}}$ & &\begin{tikzpicture}[font=\scriptsize] 
 \draw [fill=tikz_red] (0,0) rectangle (22*6/38*0.15, 0.15); 
 \draw [fill=tikz_gray] (22*6/38*0.15, 0) rectangle (6*0.15, 0.15); 
 \node [left, color=tikz_red] at (0, 0.1) {\textbf{\scriptsize{22}}\textcolor{black!50}{/38}}; 
 \end{tikzpicture} & $0.254_{\,\textcolor{black!50}{\tiny{0.05}}}$ & &\begin{tikzpicture}[font=\scriptsize] 
 \draw [fill=tikz_red] (0,0) rectangle (3*6/10*0.15, 0.15); 
 \draw [fill=tikz_gray] (3*6/10*0.15, 0) rectangle (6*0.15, 0.15); 
 \node [left, color=tikz_red] at (0, 0.1) {\textbf{\scriptsize{3}}\textcolor{black!50}{/10}}; 
 \end{tikzpicture} & $0.025_{\,\textcolor{black!50}{\tiny{0.00}}}$ \\ \texttt{M2M (M)} & \begin{tikzpicture}[font=\scriptsize] 
 \draw [fill=tikz_red] (0,0) rectangle (14*6/16*0.15, 0.15); 
 \draw [fill=tikz_gray] (14*6/16*0.15, 0) rectangle (6*0.15, 0.15); 
 \node [left, color=tikz_red] at (0, 0.1) {\textbf{\scriptsize{14}}\textcolor{black!50}{/16}}; 
 \end{tikzpicture} & $12.53_{\,\textcolor{black!50}{\tiny{1.42}}}$ & &\begin{tikzpicture}[font=\scriptsize] 
 \draw [fill=tikz_red] (0,0) rectangle (17*6/38*0.15, 0.15); 
 \draw [fill=tikz_gray] (17*6/38*0.15, 0) rectangle (6*0.15, 0.15); 
 \node [left, color=tikz_red] at (0, 0.1) {\textbf{\scriptsize{17}}\textcolor{black!50}{/38}}; 
 \end{tikzpicture} & $0.110_{\,\textcolor{black!50}{\tiny{0.00}}}$ & &\begin{tikzpicture}[font=\scriptsize] 
 \draw [fill=tikz_red] (0,0) rectangle (2*6/10*0.15, 0.15); 
 \draw [fill=tikz_gray] (2*6/10*0.15, 0) rectangle (6*0.15, 0.15); 
 \node [left, color=tikz_red] at (0, 0.1) {\textbf{\scriptsize{2}}\textcolor{black!50}{/10}}; 
 \end{tikzpicture} & $0.010_{\,\textcolor{black!50}{\tiny{0.00}}}$ \\ \texttt{M2M (L)} & \begin{tikzpicture}[font=\scriptsize] 
 \draw [fill=tikz_red] (0,0) rectangle (14*6/16*0.15, 0.15); 
 \draw [fill=tikz_gray] (14*6/16*0.15, 0) rectangle (6*0.15, 0.15); 
 \node [left, color=tikz_red] at (0, 0.1) {\textbf{\scriptsize{14}}\textcolor{black!50}{/16}}; 
 \end{tikzpicture} & $11.22_{\,\textcolor{black!50}{\tiny{2.19}}}$ & &\begin{tikzpicture}[font=\scriptsize] 
 \draw [fill=tikz_red] (0,0) rectangle (11*6/38*0.15, 0.15); 
 \draw [fill=tikz_gray] (11*6/38*0.15, 0) rectangle (6*0.15, 0.15); 
 \node [left, color=tikz_red] at (0, 0.1) {\textbf{\scriptsize{11}}\textcolor{black!50}{/38}}; 
 \end{tikzpicture} & $0.034_{\,\textcolor{black!50}{\tiny{0.00}}}$ & &\begin{tikzpicture}[font=\scriptsize] 
 \draw [fill=tikz_red] (0,0) rectangle (0*6/10*0.15, 0.15); 
 \draw [fill=tikz_gray] (0*6/10*0.15, 0) rectangle (6*0.15, 0.15); 
 \node [left, color=tikz_red] at (0, 0.1) {\textbf{\scriptsize{0}}\textcolor{black!50}{/10}}; 
 \end{tikzpicture} & $0.000_{\,\textcolor{black!50}{\tiny{0.00}}}$ \\ \bottomrule
\end{tabular}
\caption{Fraction of LPs on the English-centric setup for which models produce at least one hallucination, and average hallucination rate (and median, in subscript) across all LPs at each resource level.}
\label{tab:flores_eng_centric}
\end{table*}

\begin{figure*}
\centering
\includegraphics[width=\textwidth]{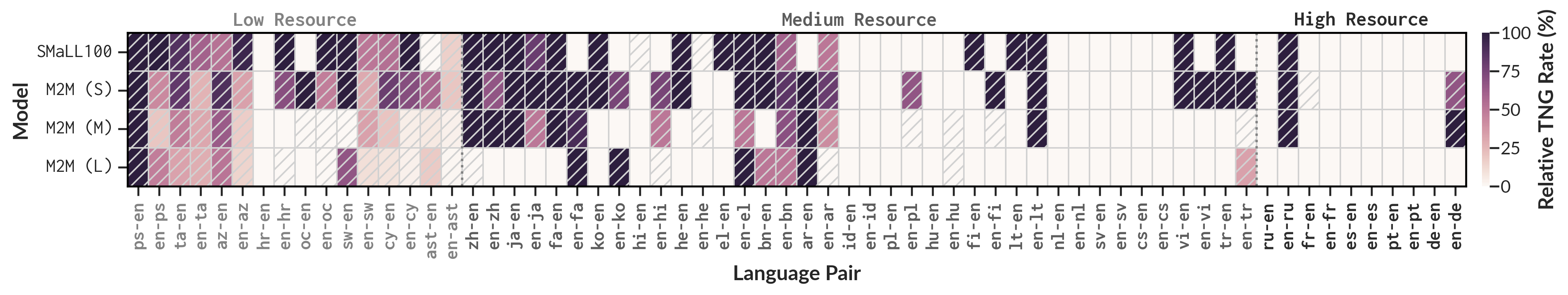}
\caption{Heatmap of the percentage of hallucinations detected with TNG (oscillatory hallucinations) among all hallucinations. Pattern-filled cells indicate at least one natural hallucination for a given model-LP combination.}
\label{fig:tng_rates_engcentric}  
\end{figure*}

\begin{table}[t]
\centering
\renewcommand\arraystretch{1}
\footnotesize
\begin{tabular}{>{\arraybackslash}m{1.35cm} >{\raggedleft\arraybackslash}m{1.5cm} >{\raggedleft\arraybackslash}m{1.5cm}}
\toprule
{\textsc{Model}} & {{\texttt{xx--en}}} & {{\texttt{en--xx}}}\\\midrule
\texttt{SMaLL100} & $0.221$ & $1.022$ \\ 
\texttt{M2M (S)} & $1.756$ & $6.152$ \\ 
\texttt{M2M (M)} & $2.290$ & $4.110$ \\ 
\texttt{M2M (L)} & $2.483$ & $3.169$ \\ \bottomrule
\end{tabular}
\caption{Average hallucination rates (\%) across all LPs at each direction (into or out of English).}
\label{tab:flores_halls_enxx_xxen}
\end{table}

\subsection{English-Centric Translation}
We start by investigating natural hallucinations on English-centric language pairs. We reveal key insights on how properties of hallucinations change across resource levels, models and translation directions. We present language-pair specific results in Appendix~\ref{app:subsec:engcentric}.

\label{subsec:analysis_eng_centric_flores}
\paragraph{Hallucinations in low-resource language pairs are not only more frequent, but also distinct.} Table~\ref{tab:flores_eng_centric} shows that hallucinations occur frequently for low-resource directions, with all M2M models exhibiting average hallucination rates exceeding 10\%. Furthermore, all models generate hallucinations for the vast majority of low-resource language pairs. On what comes to the type of hallucinations, Figure~\ref{fig:tng_rates_engcentric} demonstrates that, in contrast to mid- and high-resource language pairs, oscillatory hallucinations are less prevalent, while detached hallucinations occur more frequently in low-resource languages. This reveals that models tend to rely less on the source context when translating to or from low-resource languages. Importantly, although massive multilingual models have significantly improved translation quality for low-resource languages, these findings not only suggest that there is considerable room for improvement, but also highlight potential safety concerns arising from translations in these directions.

\begin{table*}[t]
\centering
\renewcommand\arraystretch{0.9}
\footnotesize
\begin{tabular}{>{\arraybackslash}m{1.45cm} >{\raggedleft\arraybackslash}m{1.75cm} >{\arraybackslash}m{1.15cm} c >{\raggedleft\arraybackslash}m{1.75cm} >{\arraybackslash}m{1.15cm} c >{\arraybackslash}m{1.6cm} >{\arraybackslash}m{1.15cm}}
\toprule
\multirow{2}{*}{\textsc{Model}} & \multicolumn{2}{c}{\textbf{\textsc{Low Resource}}} & & \multicolumn{2}{c}{\textbf{\textsc{Mid Resource}}} & & \multicolumn{2}{c}{\textbf{\textsc{High Resource}}}\\\cmidrule{2-3}\cmidrule{5-6}\cmidrule{8-9}
& \multicolumn{1}{l}{LP Fraction} & Rate (\%) & & \multicolumn{1}{l}{{LP Fraction}} & Rate (\%) & & LP Fraction & Rate (\%)\\\midrule
\texttt{SMaLL100} & \begin{tikzpicture}[font=\scriptsize] 
 \draw [fill=tikz_red] (0,0) rectangle (5*6/10*0.15, 0.15); 
 \draw [fill=tikz_gray] (5*6/10*0.15, 0) rectangle (6*0.15, 0.15); 
 \node [left, color=tikz_red] at (0, 0.1) {\textbf{\scriptsize{5}}\textcolor{black!50}{/10}}; 
 \end{tikzpicture} & $2.160_{\,\textcolor{black!50}{\tiny{0.02}}}$ & &\begin{tikzpicture}[font=\scriptsize] 
 \draw [fill=tikz_red] (0,0) rectangle (6*6/13*0.15, 0.15); 
 \draw [fill=tikz_gray] (6*6/13*0.15, 0) rectangle (6*0.15, 0.15); 
 \node [left, color=tikz_red] at (0, 0.1) {\textbf{\scriptsize{6}}\textcolor{black!50}{/13}}; 
 \end{tikzpicture} & $0.054_{\,\textcolor{black!50}{\tiny{0.00}}}$ & &\begin{tikzpicture}[font=\scriptsize] 
 \draw [fill=tikz_red] (0,0) rectangle (1*6/2*0.15, 0.15); 
 \draw [fill=tikz_gray] (1*6/2*0.15, 0) rectangle (6*0.15, 0.15); 
 \node [left, color=tikz_red] at (0, 0.1) {\textbf{\scriptsize{1}}\textcolor{black!50}{/2}}; 
 \end{tikzpicture} & $0.025_{\,\textcolor{black!50}{\tiny{0.02}}}$ \\ \texttt{M2M (S)} & \begin{tikzpicture}[font=\scriptsize] 
 \draw [fill=tikz_red] (0,0) rectangle (10*6/10*0.15, 0.15); 
 \draw [fill=tikz_gray] (10*6/10*0.15, 0) rectangle (6*0.15, 0.15); 
 \node [left, color=tikz_red] at (0, 0.1) {\textbf{\scriptsize{10}}\textcolor{black!50}{/10}}; 
 \end{tikzpicture} & $12.61_{\,\textcolor{black!50}{\tiny{1.79}}}$ & &\begin{tikzpicture}[font=\scriptsize] 
 \draw [fill=tikz_red] (0,0) rectangle (12*6/13*0.15, 0.15); 
 \draw [fill=tikz_gray] (12*6/13*0.15, 0) rectangle (6*0.15, 0.15); 
 \node [left, color=tikz_red] at (0, 0.1) {\textbf{\scriptsize{12}}\textcolor{black!50}{/13}}; 
 \end{tikzpicture} & $0.467_{\,\textcolor{black!50}{\tiny{0.05}}}$ & &\begin{tikzpicture}[font=\scriptsize] 
 \draw [fill=tikz_red] (0,0) rectangle (1*6/2*0.15, 0.15); 
 \draw [fill=tikz_gray] (1*6/2*0.15, 0) rectangle (6*0.15, 0.15); 
 \node [left, color=tikz_red] at (0, 0.1) {\textbf{\scriptsize{1}}\textcolor{black!50}{/2}}; 
 \end{tikzpicture} & $0.075_{\,\textcolor{black!50}{\tiny{0.07}}}$ \\ \texttt{M2M (M)} & \begin{tikzpicture}[font=\scriptsize] 
 \draw [fill=tikz_red] (0,0) rectangle (7*6/10*0.15, 0.15); 
 \draw [fill=tikz_gray] (7*6/10*0.15, 0) rectangle (6*0.15, 0.15); 
 \node [left, color=tikz_red] at (0, 0.1) {\textbf{\scriptsize{7}}\textcolor{black!50}{/10}}; 
 \end{tikzpicture} & $12.22_{\,\textcolor{black!50}{\tiny{2.41}}}$ & &\begin{tikzpicture}[font=\scriptsize] 
 \draw [fill=tikz_red] (0,0) rectangle (7*6/13*0.15, 0.15); 
 \draw [fill=tikz_gray] (7*6/13*0.15, 0) rectangle (6*0.15, 0.15); 
 \node [left, color=tikz_red] at (0, 0.1) {\textbf{\scriptsize{7}}\textcolor{black!50}{/13}}; 
 \end{tikzpicture} & $0.172_{\,\textcolor{black!50}{\tiny{0.05}}}$ & &\begin{tikzpicture}[font=\scriptsize] 
 \draw [fill=tikz_red] (0,0) rectangle (0*6/2*0.15, 0.15); 
 \draw [fill=tikz_gray] (0*6/2*0.15, 0) rectangle (6*0.15, 0.15); 
 \node [left, color=tikz_red] at (0, 0.1) {\textbf{\scriptsize{0}}\textcolor{black!50}{/2}}; 
 \end{tikzpicture} & $0.000_{\,\textcolor{black!50}{\tiny{0.00}}}$ \\ \texttt{M2M (L)} & \begin{tikzpicture}[font=\scriptsize] 
 \draw [fill=tikz_red] (0,0) rectangle (6*6/10*0.15, 0.15); 
 \draw [fill=tikz_gray] (6*6/10*0.15, 0) rectangle (6*0.15, 0.15); 
 \node [left, color=tikz_red] at (0, 0.1) {\textbf{\scriptsize{6}}\textcolor{black!50}{/10}}; 
 \end{tikzpicture} & $6.580_{\,\textcolor{black!50}{\tiny{2.02}}}$ & &\begin{tikzpicture}[font=\scriptsize] 
 \draw [fill=tikz_red] (0,0) rectangle (4*6/13*0.15, 0.15); 
 \draw [fill=tikz_gray] (4*6/13*0.15, 0) rectangle (6*0.15, 0.15); 
 \node [left, color=tikz_red] at (0, 0.1) {\textbf{\scriptsize{4}}\textcolor{black!50}{/13}}; 
 \end{tikzpicture} & $0.077_{\,\textcolor{black!50}{\tiny{0.00}}}$ & &\begin{tikzpicture}[font=\scriptsize] 
 \draw [fill=tikz_red] (0,0) rectangle (0*6/2*0.15, 0.15); 
 \draw [fill=tikz_gray] (0*6/2*0.15, 0) rectangle (6*0.15, 0.15); 
 \node [left, color=tikz_red] at (0, 0.1) {\textbf{\scriptsize{0}}\textcolor{black!50}{/2}}; 
 \end{tikzpicture} & $0.000_{\,\textcolor{black!50}{\tiny{0.00}}}$ \\ \bottomrule
\end{tabular}
\caption{Fraction of LPs on the non-English-centric setup for which models produce at least one hallucination, and average hallucination rate (and median, in subscript) across all LPs at each resource level.}
\label{tab:flores_noneng_centric}
\end{table*}

\paragraph{\texttt{SMaLL100} consistently relies more on the source text than other models.} Despite having the smallest number of parameters, \texttt{SMaLL100} shows remarkable hallucination rates across low- and mid-resource language pairs, hallucinating significantly less than its larger counterparts in low-resource settings. These improved rates may be attributed not only to the uniform sampling of language pairs during training, but also to architectural decisions. While \texttt{SMaLL100} shares a 12-layer encoder with the other models to process source representations, it diverges by employing a shallow 3-layer decoder---instead of a 12-layer decoder---and placing the target language code on the encoder side. We hypothesize that this design encourages greater reliance on the more complex encoder representations, reducing the likelihood of detachment from the source. In fact, distinct patterns in ALTI+ scores~(shown in Appendix~\ref{app:subsec:engcentric}) support this hypothesis:~\texttt{SMaLL100} consistently demonstrates higher source contributions and similar patterns across all resource levels. In contrast, M2M models show a greater tendency to rely less on the source, especially in low-resource language pairs. Importantly, however, \texttt{SMaLL100}'s reduced hallucination rates do not necessarily imply superior translation quality compared to the other M2M models: we observed a strong correlation between M2M models' corpus-level COMET-22 scores and their respective hallucination rates for low-resource languages, whereas, contrastingly, for \texttt{SMaLL100} the correlation is weak. This indicates that despite detaching less from the source content, \texttt{SMaLL100}'s translations are not necessarily of higher quality to those of other M2M models. This and other statistics can be found in the Appendix~\ref{app:subsec:engcentric}.

\paragraph{Scaling up models within the same model family leads to reduced hallucination rates.} As shown in Table~\ref{tab:flores_eng_centric}, increasing the size of the M2M family models results in consistent reductions in hallucination rates. Relative improvements are more pronounced for mid- and high-resource language pairs, with \texttt{M2M (L)} exhibiting fewer hallucinations and hallucinating for fewer languages than all other models.

\paragraph{Hallucinations are more frequent when translating out of English.} Table~\ref{tab:flores_halls_enxx_xxen} demonstrates that models are significantly more prone to hallucinate when translating out of English. In fact, in line with the observations of~\citet{ferrando-etal-2022-towards}, we found that models tend to detach more from the source text when translating out of English. This is evidenced by ALTI+ source contributions being lower across all language pairs in this direction compared to translating into English. Interestingly, we discovered that the translation direction can also influence the properties of hallucinations: (i) over 90\% of off-target hallucinations occur when translating out of English, and (ii) nearly all hallucinations into English for mid- and high-resource language pairs are oscillatory.

\begin{table*}[t]
\centering
\renewcommand\arraystretch{0.9}
\footnotesize
\begin{tabular}{>{\arraybackslash}m{1.45cm} >{\arraybackslash}m{1.25cm} >{\arraybackslash}m{1.25cm} c >{\arraybackslash}m{1.25cm} >{\arraybackslash}m{1.25cm} c >{\arraybackslash}m{1.25cm} >{\arraybackslash}m{1.25cm}}
\toprule
\multirow{2}{*}{\textsc{Model}} & \multicolumn{2}{c}{\textbf{\textsc{Low Resource}}} & & \multicolumn{2}{c}{\textbf{\textsc{Mid Resource}}} & & \multicolumn{2}{c}{\textbf{\textsc{High Resource}}}\\\cmidrule{2-3}\cmidrule{5-6}\cmidrule{8-9}
& \textsc{Flores} & TICO & & \textsc{Flores} & TICO & & \textsc{Flores} & TICO\\\midrule
\texttt{SMaLL100} & $0.448_{\,\textcolor{black!50}{\tiny{0.17}}}$ & $0.429_{\,\textcolor{black!50}{\tiny{0.29}}}$ & & $0.062_{\,\textcolor{black!50}{\tiny{0.02}}}$  & $0.083_{\,\textcolor{black!50}{\tiny{0.07}}}$ & & $0.008_{\,\textcolor{black!50}{\tiny{0.00}}}$ & $0.000_{\,\textcolor{black!50}{\tiny{0.00}}}$ \\ 
\texttt{M2M (S)} & $7.778_{\,\textcolor{black!50}{\tiny{3.48}}}$ & $6.262_{\,\textcolor{black!50}{\tiny{3.67}}}$ & & $0.087_{\,\textcolor{black!50}{\tiny{0.05}}}$ & $0.167_{\,\textcolor{black!50}{\tiny{0.17}}}$ & & $0.017_{\,\textcolor{black!50}{\tiny{0.00}}}$ & $0.024_{\,\textcolor{black!50}{\tiny{0.00}}}$ \\
\texttt{M2M (M)} & $3.484_{\,\textcolor{black!50}{\tiny{0.67}}}$ & $2.167_{\,\textcolor{black!50}{\tiny{0.33}}}$ & & $0.268_{\,\textcolor{black!50}{\tiny{0.05}}}$  & $0.363_{\,\textcolor{black!50}{\tiny{0.02}}}$ & & $0.008_{\,\textcolor{black!50}{\tiny{0.00}}}$ & $0.000_{\,\textcolor{black!50}{\tiny{0.00}}}$  \\
\texttt{M2M (L)} & $1.008_{\,\textcolor{black!50}{\tiny{0.37}}}$ & $0.596_{\,\textcolor{black!50}{\tiny{0.21}}}$ & & $0.031_{\,\textcolor{black!50}{\tiny{0.02}}}$ & $0.018_{\,\textcolor{black!50}{\tiny{0.00}}}$ & & $0.000_{\,\textcolor{black!50}{\tiny{0.00}}}$ & $0.000_{\,\textcolor{black!50}{\tiny{0.00}}}$  \\\bottomrule
\end{tabular}
\caption{Comparison between average hallucination rate (and median, in subscript) for the same LPs at each resource level for \textsc{Flores} and TICO medical data.}
\label{tab:specialized_domain_compare}
\end{table*}

\paragraph{Toxic hallucinations pose substantial safety risks.} Toxic text in translations can emerge in the form of hallucinations~\citep{nllb2022}. To assess the prevalence of toxic text in detected hallucinations, we utilized the toxicity wordlists provided by~\citet{nllb2022}. We found that toxic text primarily appears in translations out of English and almost exclusively affects low-resource language pairs. For instance, over 1 in 8 hallucinations in Tamil contain toxic text. Interestingly, these toxic hallucinations not only exhibit high lexical overlap among them, but are repeated across models for multiple unique source sentences. Moreover, they are not necessarily reduced by scaling up the model size. These observations suggest that these hallucinations are likely to be traced back to toxic patterns in the training data,\footnote{Upon inspecting the Common Crawl corpora that were used to create the training data, we found reference translations that exactly match the toxic hallucinations.} aligning with observations in~\citet{raunak-etal-2021-curious, guerreiro-etal-2022-lookingforaneedle}. Moreover, we also found that these hallucinations can be propagated through model distillation, as evidenced by \texttt{SMaLL100} generating toxic hallucinations that are copies of those of its teacher model. This underlines the necessity of rigorously filtering training data to ensure safe and responsible use of these models in real-world applications. 

\subsection{Beyond English-Centric Translation}
We shift our focus to translation directions that do not involve English, typically corresponding to directions with less supervision during training. We present language-pair specific results in Appendix~\ref{app:subsec:nonengcentric}.

\paragraph{Trends are largely similar to English-centric directions.} Table~\ref{tab:flores_noneng_centric} reveals trends that largely mirror those observed in the English-centric setup:\footnote{Comparing absolute hallucination rates between the two setups is not advised, as they involve different translation directions, which may render such comparisons unreliable.}~(i)~hallucinations are more frequent in low-resource settings; (ii)~\texttt{SMaLL100} significantly outperforms the M2M models in low-resource language pairs; and (iii)~scaling up to \texttt{M2M~(L)} consistently yields substantial improvements over the smaller M2M models in low- and mid-resource directions. Additionally, the trends related to hallucination types also hold across the two setups: detached hallucinations are more prevalent in low-resource settings, while oscillatory hallucinations overwhelmingly dominate in mid- and high-resource directions~(see Appendix~\ref{app:subsec:nonengcentric}).

\paragraph{Less supervised language pairs exhibit extremely high hallucination rates.} As expected, models struggle more with hallucinations for directions with less or even no supervision during training, such as \texttt{ro-hy} and \texttt{af-zu}. For instance, \texttt{M2M (M)} hallucinates for nearly half of the translations in these directions.

\subsection{Translation on Specialized Domains}
We now turn to investigating hallucinations in data from the medical domain, where they can have devastating consequences. Using the TICO dataset, we compare hallucination rates with the \textsc{Flores} dataset for 18 translation directions. We present language-pair specific results in Appendix~\ref{app:subsec:specialized_domain}.

\paragraph{Hallucinations are not exacerbated under medical domain data.} Table~\ref{tab:specialized_domain_compare} reveals that hallucination rates for the TICO medical data do not consistently exceed those observed for the \textsc{Flores} Wikipedia data. This finding diverges from previous works that investigated hallucinations for specialized domain data~\citep{wang-sennrich-2020-exposure, muller-etal-2020-domain}. We hypothesize that, in contrast with the smaller models typically trained on limited datasets from a single domain used in those works, the concept of "domain shift" may not be as pronounced for M2M models. These models are not only much larger but, crucially, they are trained on a dataset containing over 7 billion parallel sentences gathered from the web, which encompasses a broad array of domains. This massive training set potentially mitigates the impact of domain shift and, consequently, reduces its influence on hallucinations.

\section{Mitigation of Hallucinations through Fallback Systems}
\label{sec:mitigation}

Building upon our analysis on natural hallucinations in the previous section, we now explore the potential of reducing hallucinations and enhancing overall translation quality by employing a simple hybrid setup that can take advantage of multiple systems with possible complementary strengths. Put simply, we leverage an alternative system as a fallback when the primary original model produces hallucinations. Our analysis in the main text is focused on the more extensive English-centric setup. We provide results on the non-English-centric setup in Appendix~\ref{app:fallbacksystemanalysis}.

\subsection{Employing models of the same family as fallback systems}

We begin by analyzing the performance of same-family models when employed as fallback systems for one another (e.g., using \texttt{SMaLL100}, \texttt{M2M (M)}, and \texttt{M2M~(L)} as fallbacks for \texttt{M2M~(S)}).\footnote{For simplicity, we consider the distilled \texttt{SMaLL100} as a model from the M2M family.}

\begin{figure}
    \centering
    \includegraphics[width=\columnwidth]{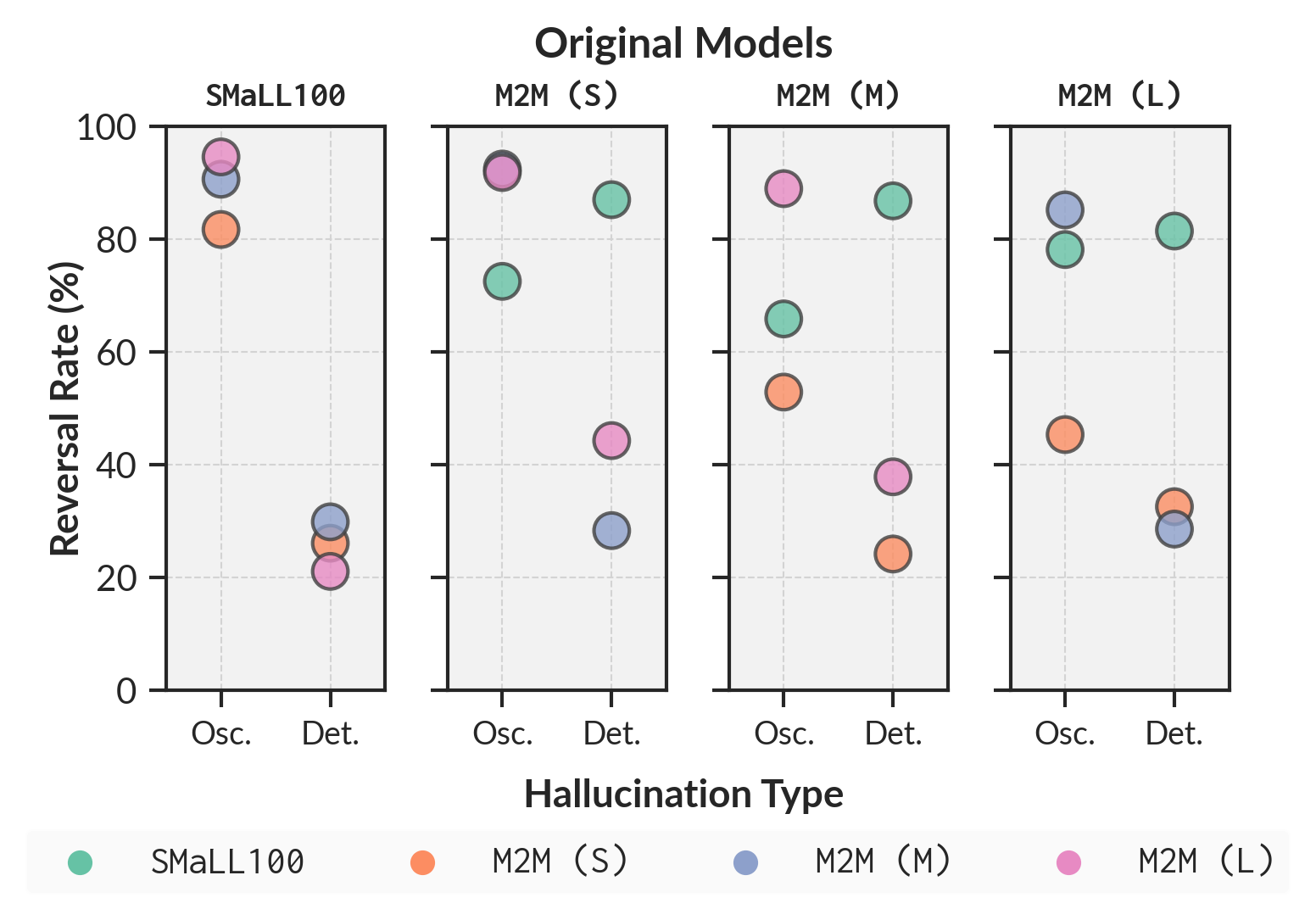}
    \caption{Reversal rates for oscillatory (\texttt{Osc.}) and detached (\texttt{Det.}) hallucinations when using models of the same family as fallback systems.}
    \label{fig:engcentric_reversal}
\end{figure}

\begin{figure*}[t]
\begin{subfigure}[b]{0.495\textwidth}
\centering
\includegraphics[scale=0.645]{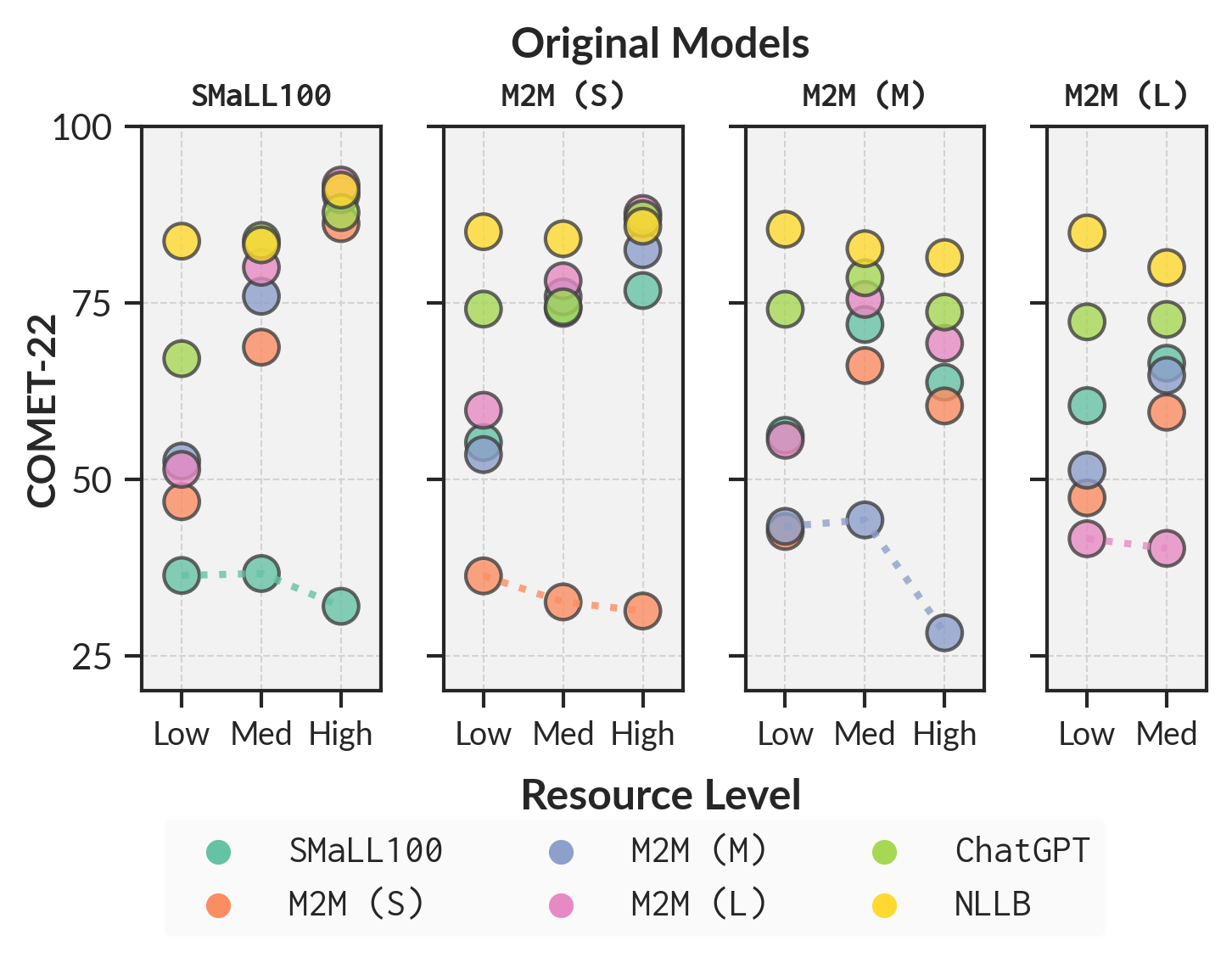}
\caption{Translation quality.}
\label{fig:flores_engcentric_fallback_quality}
\end{subfigure} \quad 
\begin{subfigure}[b]{0.495\textwidth}
\centering
\includegraphics[scale=0.555]{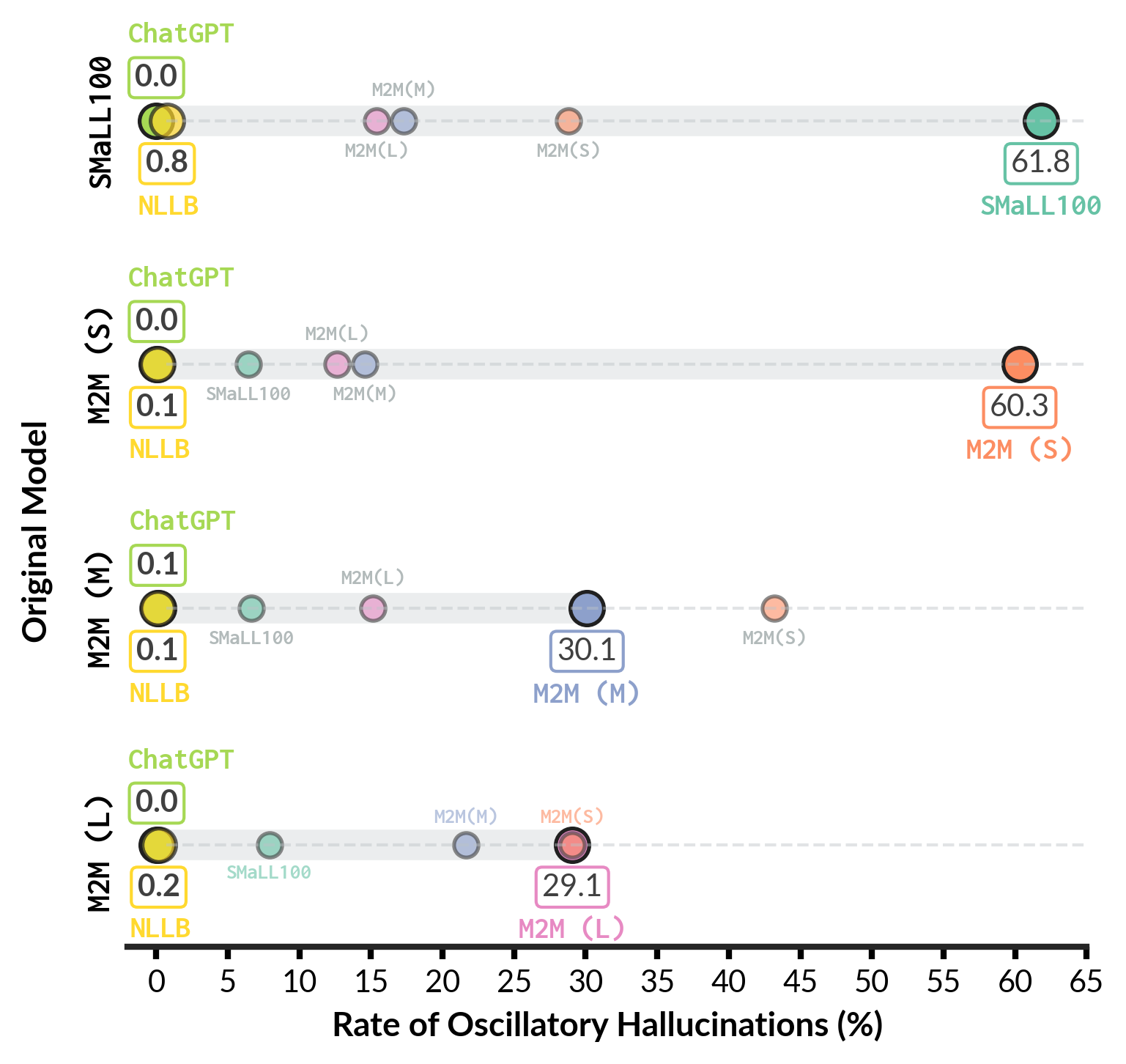}
\caption{Prevalence of oscillatory hallucinations.}
\label{fig:flores_engcentric_fallback_oscillatory}
\end{subfigure}
\caption{Fallback system analysis recurring to models of different families, such as \texttt{ChatGPT} and \texttt{NLLB}. We analyse overall translation quality improvements on the original model hallucinated translations~(represented with dashed lines) across different resource levels via COMET-22 scores in~(a), and overall prevalence of oscillatory hallucinations among the fallback translations in~(b).}
\label{fig:flores_engcentric_reversal_quality_osc}
\end{figure*}

\paragraph{Detached hallucinations are particularly \textit{sticky} across M2M models.}
Figure~\ref{fig:engcentric_reversal} reveals that when employing M2M models as fallback systems, reversal rates---percentage of hallucinations from the original system that are corrected by the fallback system---are consistently higher for oscillatory hallucinations than for detached hallucinations. These findings not only corroborate those in~\citet{guerreiro-etal-2022-lookingforaneedle}, where oscillatory hallucinations were found to be less related to model defects, but also further emphasize the close connection between detached hallucinations and training data. This connection can help explain their \textit{stickiness}: since the M2M models share the same training data, reversing these hallucinations using other same-family models as fallbacks is more challenging. Interestingly, we also observe that \texttt{M2M (L)} particularly struggles to reverse the detached hallucinations generated by its distilled counterpart \texttt{SMaLL100}, suggesting that model defects can persist and be shared during distillation.

\paragraph{Scaling up within the model family is not an effective strategy for mitigating hallucinations.} In line with our analysis in Section~\ref{subsec:analysis_eng_centric_flores}, Figure~\ref{fig:engcentric_reversal} shows that reversal rates using \texttt{SMaLL100} as a fallback system are higher for detached hallucinations than for oscillatory hallucinations. In fact, although \texttt{SMaLL100} is a distilled M2M model, its training data, training procedure, and architecture differ from those of the M2M models. This distinction may make it more complementary as a fallback system to other M2M models than simply scaling up within the same model family. This suggests that merely increasing the scale of models within the same family is not an effective strategy for mitigating hallucinations, and exploring alternative models with different architectures and trained on different data could yield more substantial improvements. In the next section, we will analyze this alternative strategy.

\subsection{Employing external models as fallback systems}

Motivated by the findings from the previous section, we will now study how models that are not from the M2M family can be employed to further mitigate hallucinations and improve translation quality. We will test this approach with two different models: (i)~we will prompt \texttt{ChatGPT} as detailed in Section~\ref{sec:experimental_suite}, and (ii)~we will use a high-quality 3.3B parameter model from the NLLB family of multilingual NMT models~(\texttt{NLLB}) proposed in~\citet{nllb2022}.

\paragraph{Translation quality can be significantly improved with external fallback systems.} Figure~\ref{fig:flores_engcentric_fallback_quality} demonstrates that external fallback systems, particularly \texttt{NLLB}, can significantly enhance translation quality of originally hallucinated translations compared to same-family models.\footnote{For reference, in the English-centric setup, the averaged COMET-22 corpus-level scores for low-, mid-, and high-resource LPs obtained with \texttt{M2M(L)} are $73.63$, $86.54$, and $87.19$, respectively. We provide fallback system quality scores for all language pairs in Appendix~\ref{app:fallbacksystemanalysis}.} This improvement is especially notable for low-resource languages, where both \texttt{ChatGPT} and \texttt{NLLB} consistently boost translation quality. Remarkably, \texttt{NLLB} generally outperforms \texttt{ChatGPT} as a fallback system for low- and mid-resource languages, aligning with the findings in~\citet{hendyetal2023_microsoftchatgpt}, which revealed that GPT models have limited capabilities for lower-resourced languages and lag behind dedicated translation models in those settings. Nonetheless, \texttt{ChatGPT} still surpasses dedicated M2M translation systems in these resource levels when used as a fallback system, underscoring the limitations of relying on same-family models as fallback systems. 

\paragraph{Oscillatory hallucinations are practically non-existent when \texttt{ChatGPT} is the fallback system.} From Figure~\ref{fig:flores_engcentric_fallback_oscillatory}, we see another benefit of employing external fallback systems: oscillatory hallucinations are almost entirely eliminated. Interestingly, consistent with our findings in Section~\ref{sec:halls_under_perturbation}, we observe that \texttt{ChatGPT} produces very few, if any, oscillations, slightly improving the rates obtained with \texttt{NLLB}. This provides further evidence that, although hallucinations obtained via prompting LLMs may still occur, they exhibit different properties and surface forms. Investigating and understanding these differences in hallucination properties presents an interesting research path for future work.

\section{Conclusion}
\label{sec:conclusions}
We have comprehensively investigated the phenomenon of hallucinations in massively multilingual translation models. By departing from the settings studied in previous work that focused on bilingual models trained on high-resource language pairs, we were able to explore a wide range of research scenarios that remained overlooked. 

Our analysis revealed several key insights on the prevalence and properties of hallucinations across various models of different scale, translation directions, and data conditions, including: the prevalence of hallucinations across multiple translation directions across different resource levels and beyond English-centric translation; the emergence of toxicity in hallucinations; and the effect of scaling up within the same model family on the prevalence of hallucinations. Additionally, we explored how fallback systems can mitigate hallucinations and improve overall translation quality. We found that hallucinations can be \textit{sticky} and difficult to reverse when using models that share the same training data and architecture. However, by leveraging other external models, we can significantly improve translation performance and virtually eliminate pathologies such as oscillatory hallucinations.

To support future research on this topic, we are open-sourcing our code and releasing over a million translations and detection results across several models and language pairs.


\section*{Limitations}
Our study mainly focuses on the M2M family of multilingual models. We chose this family of models as it includes several models at different sizes and the largest open-source multilingual NMT model. It is unclear how our findings generalize to other families of multilingual models~(e.g., the NLLB family of models).

Our detection approaches inherit the limitations that carry over with the metrics that are leveraged in them. For instance, following all of previous work, we adopt a BLEU metric to detect hallucinations under perturbation. However, this and other lexical metrics ranked worst than reference-based neural metrics in last year's WMT22 Metrics Shared Task~\citep{freitag-etal-2022-results}. 

We analyzed \texttt{ChatGPT} as it has demonstrated impressive capabilities for translation and other multilingual tasks, such as MT evaluation. Unfortunately, the model remains behind API walls and documentation is scarce. As such, we could not ensure that \texttt{ChatGPT} was not trained on our evaluation sets, nor could we evaluate the contribution of the source text to \texttt{ChatGPT}'s translations, which would have enabled detection of detached hallucinations. Despite these limitations, we believe our findings provide relevant insights into the properties of translations generated by the model.

\section*{Acknowledgments}
We would like to thank Meta AI for open-sourcing the M2M models and maintaining libraries such as \texttt{stopes}~\citep{andrews-etal-2022-stopes} and \texttt{nllb}~\citep{nllb2022}. The work is partially supported by the European Research Council (ERC StG DeepSPIN 758969), by EU’s Horizon Europe Research and Innovation Actions (UTTER, contract 101070631), by the FCT through contract UIDB/50008/2020, and by the projects MAIA and NextGenAI (LISBOA-01-0247-FEDER-045909 and 2022-C05i0102-02). Part of this work was performed using HPC resources from GENCI-IDRIS (Grant 2022- AD01101838).

\newpage

\bibliography{tacl2021, new_anthology, custom}
\bibliographystyle{acl_natbib}

\onecolumn
\appendix 
\onecolumn
\begin{center}
    \Large{\textbf{Supplemental Material}}
\end{center}
\section{List of Languages}
We summarize the languages used in our work in Table~\ref{tab:app:full_list_of_languages}.\footnote{We determine the resource levels following~\citet{goyal-etal-2022-flores}. We consider very-low resource languages as low-resource.}
\begin{table}[H]
\centering
\renewcommand\arraystretch{0.915}
\footnotesize
\begin{tabular}{lllll}
\toprule
\textbf{ISO} & \textbf{Language} & \textbf{Language grouping} & \textbf{Script} & \textbf{Resource Level}\\
\midrule
\texttt{af} & Afrikaans & Germanic & Latin & Low \\
\texttt{ar} & \textbf{Arabic} & Arabic & Arabic & Medium \\
\texttt{ast} & Asturian & Romance & Latin & Low \\
\texttt{az} & Azerbaijani & Turkic & Latin & Low \\
\texttt{be} & Belarusian & Slavic & Cyrillic & Low \\
\texttt{bn} & \textbf{Bengali} & Indo-Aryan & Bengali & Medium \\
\texttt{bs} & Bosnian & Slavic & Latin & Low \\
\texttt{ca} & Catalan & Romance & Latin & Medium \\
\texttt{cs} & \textbf{Czech} & Slavic & Latin & Medium \\
\texttt{cy} & Welsh & Celtic & Latin & Low \\
\texttt{de} & \textbf{German} & Germanic & Latin & High \\
\texttt{el} & \textbf{Greek} & Hellenic & Greek & Medium \\
\texttt{en} & \textbf{English} & Germanic & Latin & High \\
\texttt{es} & \textbf{Spanish} & Romance & Latin & High \\
\texttt{et} & Estonian & Uralic & Latin & Medium \\
\texttt{fa} & \textbf{Persian} & Iranian & Arabic & Medium \\
\texttt{fi} & \textbf{Finnish} & Uralic & Latin & Medium \\
\texttt{fr} & \textbf{French} & Romance & Latin & High \\
\texttt{gu} & Gujarati & Indo-Aryan & Gujarati & Low \\
\texttt{ha} & Hausa & Afro-Asiatic & Latin & Low \\
\texttt{he} & \textbf{Hebrew} & Semitic & Hebrew & Medium \\
\texttt{hi} & \textbf{Hindi} & Indo-Aryan & Devanagari & Medium \\
\texttt{hr} & Croatian & Slavic & Latin & Low \\
\texttt{hu} & \textbf{Hungarian} & Uralic & Latin & Medium \\
\texttt{hy} & Armenian & Armenian & Armenian & Low \\
\texttt{id} & \textbf{Indonesian} & Malayo-Polynesian & Latin & Medium \\
\texttt{it} & Italian & Romance & Latin & High \\
\texttt{is} & Icelandic & Germanic & Latin & Medium \\
\texttt{ja} & \textbf{Japanese} & Japonic & Kanji; Kana & Medium \\
\texttt{kk} & Kazakh & Turkic & Cyrillic & Low \\
\texttt{km} & Central Khmer & Khmer & Khmer & Low \\
\texttt{ko} & \textbf{Korean} & Koreanic & Hangul & Medium \\
\texttt{lt} & \textbf{Lithuanian} & Baltic & Latin & Medium \\
\texttt{lv} & Latvian & Baltic & Latin & Medium \\
\texttt{mk} & Macedonian & Slavic & Cyrillic & Medium \\
\texttt{mr} & Marathi & Indo-Aryan & Devanagari & Low \\
\texttt{nl} & \textbf{Dutch} & Germanic & Latin & Medium \\
\texttt{oc} & Occitan & Romance & Latin & Low \\
\texttt{pl} & \textbf{Polish} & Slavic & Latin & Medium \\
\texttt{ps} & Pashto & Iranian & Arabic & Low \\
\texttt{pt} & \textbf{Portuguese} & Romance & Latin & High \\
\texttt{ro} & Romanian & Romance & Latin & Medium \\
\texttt{ru} & \textbf{Russian} & Slavic & Cyrillic & High \\
\texttt{sk} & Slovak & Slavic & Latin & Medium \\
\texttt{sr} & Serbian & Slavic & Cyrillic; Latin & Medium \\
\texttt{sv} & \textbf{Swedish} & Germanic & Latin & Medium\\
\texttt{sw} & \textbf{Swahili} & Niger-Congo & Latin & Low \\
\texttt{ta} & \textbf{Tamil} & Dravidian & Tamil & Low \\
\texttt{tl} & Tagalog & Malayo-Polynesian & Latin & Low \\
\texttt{tr} & \textbf{Turkish} & Turkic & Latin & Medium\\
\texttt{uk} & Ukrainian & Slavic & Cyrillic & Medium\\
\texttt{vi} & \textbf{Vietnamese} & Vietic & Latin & Medium\\
\texttt{zh} & \textbf{Chinese Mandarin} & Chinese & Chinese & Medium\\
\texttt{zu} & Zulu & Niger-Congo & Latin & Low\\ \bottomrule
\end{tabular}
\caption{List of languages that we use in our work. We fill the table with the information available in~\citet{fanetal2020_m2m}. For each language, we display the ISO code, language name, language grouping, script and resource level. The languages in \textbf{bold} are the bridge languages.}
\label{tab:app:full_list_of_languages}
\end{table}

\section{Dataset Statistics}
\label{app:datasets}

\paragraph{\textsc{Flores}.}  The \textsc{Flores} benchmark~\citep{goyal-etal-2022-flores} consists of a multi-parallel dataset in 101 languages with sentences extracted from Wikipedia. The sentences in the dataset are translated from original English sentences. We join the \texttt{dev} and \texttt{devtest} splits for a total of 2009 parallel sentences for each translation direction.

\paragraph{TICO.} The TICO benchmark~\citep{anastasopoulos-etal-2020-tico} is a multilingual dataset specialized in the medical domain obtained by combining English open-source data from various sources, such as scientific articles, government health announcements, and others. We join the \texttt{dev} and \texttt{test} set for a total of 2100 parallel sentences for each translation direction.

\paragraph{WMT benchmarks.} We selected the same English-centric WMT benchmarks that were used as evaluation sets in the original M2M paper~\citep{fanetal2020_m2m}. Additionally, we selected recent WMT test sets from the WMT21 and WMT22 campaigns as they were released after the models were trained. A summary of the datasets used for each translation direction can be found in Table~\ref{app:wmtbenchmarks_ref}.

\begin{table}[H]
\renewcommand\arraystretch{0.975}
\centering
\footnotesize
\begin{tabular}{cccccccccccccc}
\toprule
\texttt{cs} & \texttt{de} & \texttt{et} & \texttt{fi} & \texttt{fr} & \texttt{ha} & \texttt{is} & \texttt{ja} & \texttt{lt} & \texttt{lv} & \texttt{ru} & \texttt{tr} & \texttt{uk} & \texttt{zh} \\
\midrule
2018 & 2019 & 2018 & 2017 & 2014 & 2021 & 2021 & 2022 & 2019 & 2017 & 2019 & 2017 & 2022 & 2019 \\
\bottomrule
\end{tabular}
\caption{Specification of the WMT benchmarks used in this work.}
\label{app:wmtbenchmarks_ref}
\end{table}

\section{Hallucinations under Perturbation}
\subsection{Data.}
\label{app:data_for_halls_under_perturbation}

\paragraph{Perturbations.} We randomly misspell words by changing the characters with a probability of 0.01;\footnote{We use the \texttt{ButterFingersPerturbation} from the NL-Augmenter framework~\citep{dhole2022nlaugmenter}.} insert a token chosen randomly from the 50 most frequent tokens or punctuation in the test set; and, randomly title-case words with a probability of 0.1, guaranteeing that at least one word's capitalization is perturbed.

\paragraph{Languages.} We pair the following languages with English: \texttt{af}, \texttt{ar}, \texttt{ast}, \texttt{bn}, \texttt{cs}, \texttt{cy}, \texttt{de}, \texttt{el}, \texttt{es}, \texttt{fa}, \texttt{fi}, \texttt{fr}, \texttt{he}, \texttt{hi}, \texttt{hr}, \texttt{hu}, \texttt{id}, \texttt{ja}, \texttt{ko}, \texttt{lt}, \texttt{nl}, \texttt{oc}, \texttt{pl}, \texttt{pt}, \texttt{ru}, \texttt{sv}, \texttt{sw}, \texttt{tl}, \texttt{tr}, \texttt{vi}, \texttt{zh}. We selected all bridge languages~(see Table~\ref{tab:app:full_list_of_languages}) and additional low-resource languages, as they are underrepresented among the bridge languages. Although Tamil is a bridge language, we did not include it in our analysis, as we could not find enough candidates with reasonable quality for hallucinations under perturbation after applying rule (i) of the detection approach~(see Section~\ref{subsec:halls_under_perturb_eval_setting}). 

\subsection{Supplementary Results}
\label{app:supp_results_for_hall_under_perturbation}
\paragraph{Correlation between original translation quality and hallucination under perturbation.} We present the statistics in Table~\ref{app:halls_under_perturb_correlation}. As mentioned in the main text, the Pearson correlation between the original translation quality and the detection outputs of hallucinations under perturbation is very weak across all models. 

\begin{table}[h]
\footnotesize
\renewcommand\arraystretch{0.975}
\centering
\begin{tabular}{rrrr}
\toprule
\texttt{SMaLL100} & \texttt{M2M (S)} & \texttt{M2M (M)} & \texttt{M2M (L)} \\\midrule
$-0.03$ & $-0.05$ & $-0.03$ & $-0.02$\\
\bottomrule
\end{tabular}
\caption{Model-based Pearson correlation scores between original spBLEU scores and hallucination under perturbation assignments across all language pairs.}
\label{app:halls_under_perturb_correlation}
\end{table}

\paragraph{Discussion on hallucinations under perturbation and natural hallucinations.} As we have mentioned in the main text, hallucinations under perturbation, in contrast to natural hallucinations, explicitly reveal the lack of robustness of translation systems. Put simply, hallucinations under perturbation are cases where the model is not robust and, as a result, undergoes significant negative shifts in translation quality \textit{due to perturbations}. This means that they do not necessarily entail detachment from the source. Hallucinations under perturbation have been studied as such in previous work, and detection of these translations with the 2-step method that we adopted in our work~(see Section~\ref{subsec:halls_under_perturb_eval_setting}) has been consistently used in all of previous research. We decided to keep the same designations, definitions and detection approach so as to be consistent with previous work. Nevertheless, we assess here whether the detected hallucinations under perturbation are, in fact, detached or oscillatory hallucinations. We found that more than 85\% of hallucinations under perturbation would be detected with our detection approach for natural hallucinations. Inspection of the non-detected translations revealed that they usually contain critical mistranslation errors, but do not necessarily exhibit detachment from the source content (e.g., off-target translations or copies of the source).

\begin{figure}[H]
    \footnotesize
    \centering
   \includegraphics[width=\textwidth]{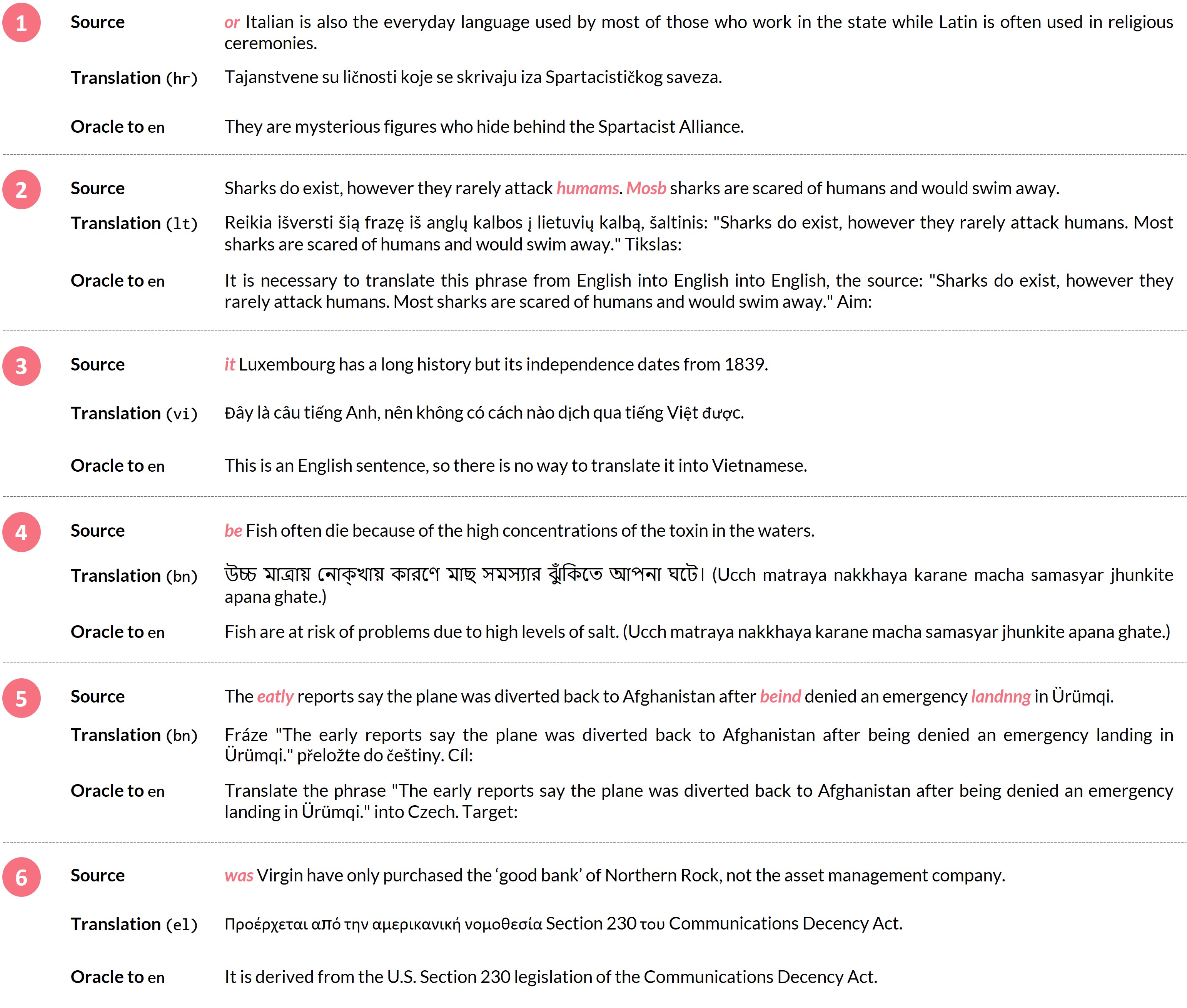}
    \caption{Examples of hallucinations under perturbation generated by \texttt{ChatGPT}. We present an oracle English translation of the generated hallucination with Bing Microsoft Translator (\textsc{Oracle to \texttt{en}}). The source perturbations are shown in \textit{italic} and coloured differently to the rest of the text.
} 
    \label{fig:app:chatgpt_halls}
\end{figure}

\paragraph{Examples of hallucinations under perturbation generated by \texttt{ChatGPT}.} We provide examples of hallucinations under perturbation in Figure~\ref{fig:app:chatgpt_halls}. 

\section{Natural Hallucinations}
Hereinafter, we will use the terms natural hallucinations and hallucinations interchangeably.
\subsection{Translation directions.}

\paragraph{English-Centric setup.} We pair the following languages with English: \texttt{ar}, \texttt{ast}, \texttt{az}, \texttt{bn}, \texttt{cs}, \texttt{cy}, \texttt{de}, \texttt{el}, \texttt{es}, \texttt{fa}, \texttt{fi}, \texttt{fr}, \texttt{he}, \texttt{hi}, \texttt{hr}, \texttt{hu}, \texttt{id}, \texttt{ja}, \texttt{ko}, \texttt{lt}, \texttt{nl}, \texttt{oc}, \texttt{pl}, \texttt{ps}, \texttt{pt}, \texttt{ru}, \texttt{sv}, \texttt{sw}, \texttt{ta}, \texttt{tr}, \texttt{vi}, \texttt{zh}. Similarly to the choice for the experiments on hallucinations under perturbation, we selected all the bridge languages and additional low-resource languages.

\paragraph{Non-English-Centric setup.} We analyzed the following translation directions: \texttt{hi-bn}, \texttt{it-fr}, \texttt{de-hu}, \texttt{it-de}, \texttt{cs-sk}, \texttt{nl-fr}, \texttt{fr-sw}, \texttt{ro-ru}, \texttt{ro-uk}, \texttt{de-hr}, \texttt{hr-sr}, \texttt{be-ru}, \texttt{hr-hu}, \texttt{hr-cs}, \texttt{el-tr}, \texttt{hr-sk}, \texttt{nl-de}, \texttt{af-zu}, \texttt{ro-hu}, \texttt{hi-mr}, \texttt{ro-tr}, \texttt{uk-ru}, \texttt{ro-hy}, \texttt{ar-fr}, \texttt{ro-de}. We selected language pairs that were used in the analysis from the original M2M paper on real-world settings for many-to-many translation~\citep{fanetal2020_m2m}. They represent translation directions that are commonly used in countries that have official and regional languages that do not include English.

\paragraph{Specialized Domains setup.} We pair the following languages with English: \texttt{ar}, \texttt{fr}, \texttt{hi}, \texttt{id}, \texttt{ps}, \texttt{pt}, \texttt{ru}, \texttt{sw}, \texttt{zh}. We selected languages from the TICO benchmark that are supported by the M2M models.

\label{app:subsec_translation_directions}

\subsection{English-Centric Translation}
\label{app:subsec:engcentric}
\subsubsection{Evaluation with \textsc{Flores}}
\paragraph{Evaluation scores.} We present spBLEU and COMET-22 corpus-level scores for the \textsc{Flores} setup in Table~\ref{app:tab:quality_flores_setup_engcentric}.

\begin{sidewaysfigure}
    \centering
    \includegraphics[width=\textwidth]{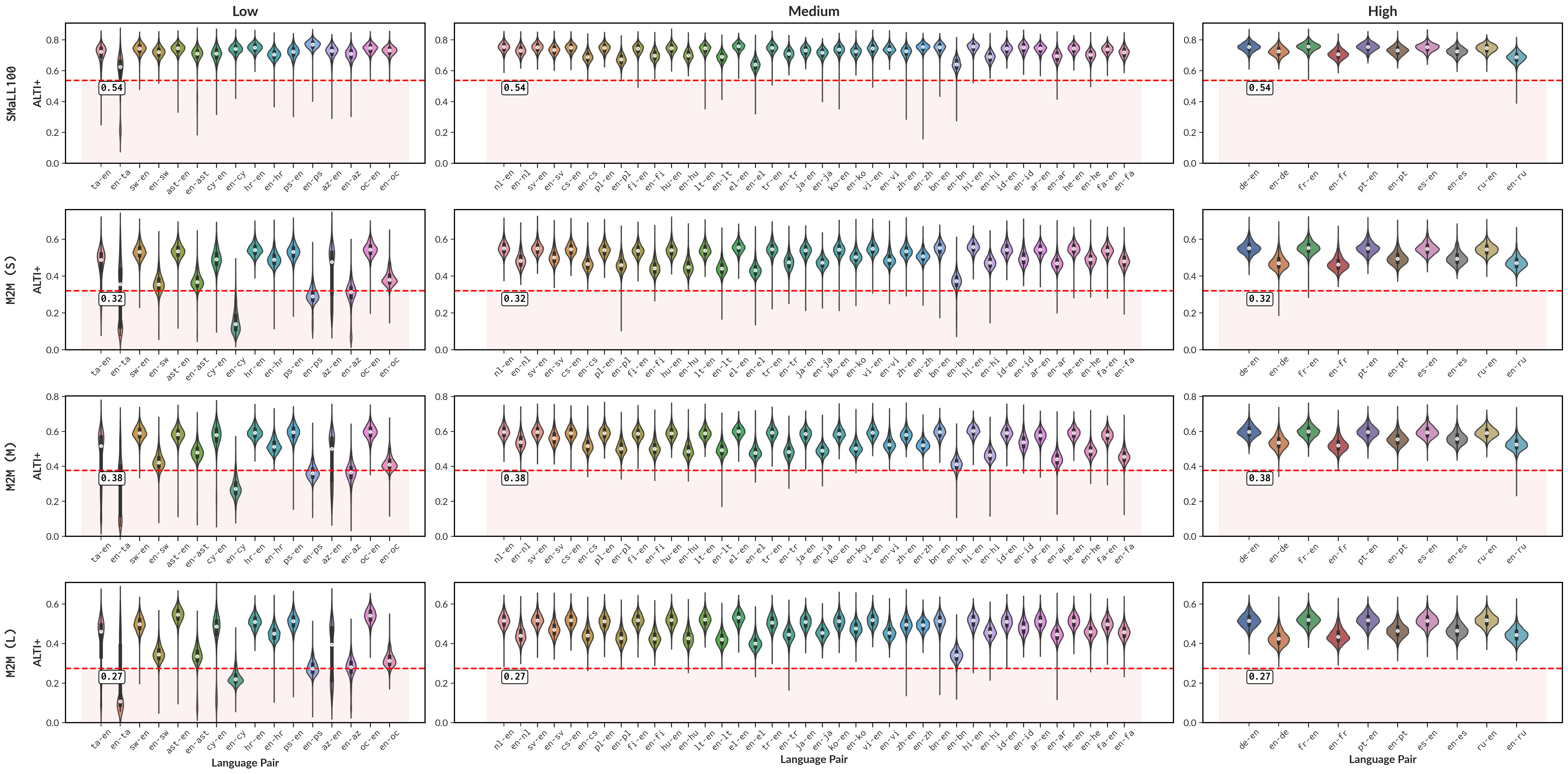}
    \caption{ALTI+ scores for all translation directions for each of the models considered in the English-centric \textsc{Flores} setup. We represent with horizontal red lines the model-based specific thresholds tuned on the validation data.}
    \label{fig:alti_flores_engcentric}
\end{sidewaysfigure}

\paragraph{ALTI+ scores.} We present the distribution of ALTI+ scores for all translation directions in Figure~\ref{fig:alti_flores_engcentric}. For the M2M models, we find the same trends presented in~\citet{ferrando-etal-2022-towards}: smaller source contributions when translating out of English and for low-resource language pairs. Remarkably, it is also clear from the distributions that ALTI+ scores for \texttt{SMaLL100} are not only significantly higher than those of the M2M models, but they are also more consistent across different resource levels.

\paragraph{Correlation between COMET-22 scores and hallucination rates.} We present the statistics in Table~\ref{app:tab:correlations_comet22_halls_rates}. As we have remarked in the main text, \texttt{SMaLL100}'s reduced hallucination rates for low-resource language pairs do not necessarily equate to superior translation quality compared to the other M2M models. We observed a strong correlation between M2M models' corpus-level COMET-22 scores and their respective hallucination rates for low-resource languages, whereas, contrastingly, for \texttt{SMaLL100} the correlation is weak. This indicates that despite detaching less from the source content, \texttt{SMaLL100}'s translations are not necessarily of higher quality to those of other M2M models. Interestingly, the correlations for the M2M models are significantly weakened on high-resource language pairs, mirroring observations from previous work~\citep{Lee2018HallucinationsIN}.

\begin{table}[H]
\renewcommand\arraystretch{0.975}
\scriptsize
\centering
\begin{tabular}{lrcrcrcr}
\toprule
\multirow{2}{*}{\textsc{Model}} &  \multicolumn{7}{c}{\textsc{Resource Level}}\\\cmidrule{2-8}
& Low & & Medium & & High & All \\\midrule
\texttt{SMaLL100} & $-0.39$ & & $-0.14$ & & $-0.27$ & $-0.52$\\
\texttt{M2M (S)} & $-0.82$ & & $-0.67$ & & $-0.49$ & $-0.81$\\
\texttt{M2M (M)} & $-0.80$ & & $-0.68$ & & $-0.15$ & $-0.79$\\
\texttt{M2M (L)} & $-0.85$ & & $-0.38$ & & --- & $-0.83$\\
\bottomrule
\end{tabular}
\caption{Model-based average Pearson correlation scores between COMET-22 and hallucination rates across all language pairs at each resource level.}
\label{app:tab:correlations_comet22_halls_rates}
\end{table}

\begin{table}[H]
\renewcommand\arraystretch{0.95}
\scriptsize
\centering
\begin{tabular}{lrrcrrcrrcrr}
\toprule
\multirow{2}{*}{LP} &  \multicolumn{2}{c}{\texttt{SMaLL100}} & &  \multicolumn{2}{c}{\texttt{M2M (S)}} & & \multicolumn{2}{c}{\texttt{M2M (M)}} & & \multicolumn{2}{c}{\texttt{M2M (L)}}\\\cmidrule{2-3}\cmidrule{5-6}\cmidrule{8-9}\cmidrule{11-12}
& spBLEU &  COMET-22 & &  spBLEU &  COMET-22 & &  spBLEU &  COMET-22 & &  spBLEU &  COMET-22 \\
\midrule
\multicolumn{12}{c}{\texttt{en-xx} \textit{directions}}\\ \cdashlinelr{1-12}
    \texttt{en-ar} &        25.96 &        81.01  & &        25.79 &        81.27  & &        16.62 &        74.92  & &        29.74 &        84.43 \\
   \texttt{en-ast} &        25.90 &        67.26  & &        24.79 &        67.80  & &        32.99 &        69.84  & &        30.68 &        67.58 \\
    \texttt{en-az} &         8.09 &        69.48  & &         4.15 &        63.38  & &         7.90 &        71.35  & &        10.18 &        76.12 \\
    \texttt{en-bn} &        27.06 &        80.70  & &        16.48 &        71.01  & &        25.53 &        81.78  & &        28.87 &        82.89 \\
    \texttt{en-cs} &        31.02 &        84.02  & &        30.52 &        83.58  & &        37.06 &        88.35  & &        37.79 &        88.74 \\
    \texttt{en-cy} &         9.44 &        46.41  & &         3.27 &        34.13  & &        14.34 &        56.19  & &        26.54 &        71.24 \\
    \texttt{en-de} &        33.77 &        81.06  & &        30.99 &        79.40  & &        39.64 &        85.57  & &        41.76 &        86.31 \\
    \texttt{en-el} &        27.55 &        84.66  & &        26.70 &        84.19  & &        32.65 &        87.98  & &        33.92 &        88.33 \\
    \texttt{en-es} &        26.55 &        82.15  & &        25.36 &        81.07  & &        29.08 &        84.42  & &        29.98 &        85.13 \\
    \texttt{en-fa} &        27.02 &        80.51  & &        27.53 &        81.47  & &        22.69 &        78.56  & &        29.83 &        84.32 \\
    \texttt{en-fi} &        23.93 &        84.31  & &        22.33 &        83.93  & &        29.41 &        89.04  & &        31.59 &        89.82 \\
    \texttt{en-fr} &        43.33 &        82.57  & &        42.30 &        81.99  & &        49.16 &        86.26  & &        51.58 &        86.94 \\
    \texttt{en-he} &        28.06 &        80.55  & &        27.09 &        80.79  & &        30.88 &        83.47  & &        34.58 &        85.23 \\
    \texttt{en-hi} &        32.91 &        74.38  & &        31.72 &        74.44  & &        32.61 &        75.79  & &        36.29 &        77.72 \\
    \texttt{en-hr} &        29.03 &        84.92  & &        27.26 &        84.21  & &        32.78 &        88.37  & &        33.65 &        88.76 \\
    \texttt{en-hu} &        27.00 &        81.29  & &        26.35 &        81.32  & &        32.54 &        86.72  & &        34.06 &        87.15 \\
    \texttt{en-id} &        39.55 &        87.68  & &        36.45 &        87.05  & &        42.50 &        89.46  & &        44.90 &        90.06 \\
    \texttt{en-ja} &        22.37 &        84.38  & &        23.70 &        84.80  & &        27.61 &        87.61  & &        28.76 &        88.18 \\
    \texttt{en-ko} &        19.36 &        82.89  & &        19.71 &        83.11  & &        22.07 &        85.46  & &        22.99 &        86.20 \\
    \texttt{en-lt} &        28.02 &        82.73  & &        26.03 &        82.54  & &        32.74 &        87.41  & &        33.71 &        87.61 \\
    \texttt{en-nl} &        27.66 &        82.73  & &        25.94 &        81.25  & &        31.01 &        85.51  & &        32.56 &        86.39 \\
    \texttt{en-oc} &        30.40 &        68.86  & &        23.27 &        67.95  & &        30.07 &        70.29  & &        37.25 &        70.39 \\
    \texttt{en-pl} &        21.67 &        81.56  & &        21.18 &        81.59  & &        25.51 &        86.26  & &        26.38 &        86.92 \\
    \texttt{en-ps} &         6.56 &        58.05  & &         5.79 &        55.60  & &         9.26 &        62.55  & &        13.22 &        68.90 \\
    \texttt{en-pt} &        44.45 &        85.92  & &        43.58 &        85.35  & &        49.53 &        88.29  & &        51.54 &        88.77 \\
    \texttt{en-ru} &        27.84 &        82.32  & &        26.94 &        81.83  & &        33.41 &        86.61  & &        35.15 &        87.77 \\
    \texttt{en-sv} &        40.65 &        86.05  & &        39.45 &        85.40  & &        46.37 &        88.95  & &        47.52 &        89.10 \\
    \texttt{en-sw} &        32.22 &        79.73  & &        21.42 &        72.98  & &        30.27 &        80.33  & &        35.21 &        81.67 \\
    \texttt{en-ta} &         9.73 &        62.13  & &         5.45 &        54.28  & &         5.12 &        56.82  & &         3.87 &        50.11 \\
    \texttt{en-tr} &        28.59 &        83.48  & &        28.90 &        84.18  & &        30.63 &        86.49  & &        33.26 &        87.42 \\
    \texttt{en-vi} &        35.87 &        83.74  & &        34.32 &        82.48  & &        39.51 &        86.54  & &        41.03 &        87.02 \\
    \texttt{en-zh} &        18.47 &        78.61  & &        19.12 &        78.73  & &        23.11 &        83.11  & &        22.75 &        82.92 \\\midrule
    
    \multicolumn{12}{c}{\texttt{xx-en} \textit{directions}}\\ \cdashlinelr{1-12}
    \texttt{ar-en} &        30.92 &        82.48  & &        29.92 &        82.14  & &        29.58 &        80.81  & &        36.79 &        85.39 \\
   \texttt{ast-en} &        34.45 &        77.59  & &        30.34 &        74.49  & &        36.78 &        79.80  & &        38.71 &        80.93 \\
    \texttt{az-en} &        15.38 &        75.35  & &         7.35 &        58.69  & &         9.19 &        61.38  & &         9.51 &        61.09 \\
    \texttt{bn-en} &        25.96 &        83.54  & &        25.64 &        83.33  & &        28.40 &        84.51  & &        31.43 &        86.03 \\
    \texttt{cs-en} &        34.34 &        84.92  & &        33.78 &        84.45  & &        38.71 &        86.88  & &        40.74 &        87.58 \\
    \texttt{cy-en} &        19.64 &        57.77  & &        13.06 &        51.64  & &        27.86 &        67.19  & &        30.49 &        69.31 \\
    \texttt{de-en} &        38.21 &        85.61  & &        36.89 &        84.87  & &        42.85 &        87.77  & &        44.75 &        88.30 \\
    \texttt{el-en} &        30.59 &        84.18  & &        29.09 &        83.23  & &        35.39 &        86.45  & &        37.03 &        87.13 \\
    \texttt{es-en} &        27.61 &        83.57  & &        25.75 &        82.86  & &        29.54 &        85.22  & &        31.22 &        85.92 \\
    \texttt{fa-en} &        28.97 &        83.18  & &        28.17 &        82.73  & &        30.03 &        83.37  & &        34.28 &        85.99 \\
    \texttt{fi-en} &        29.09 &        85.50  & &        27.95 &        84.49  & &        33.81 &        88.07  & &        35.35 &        88.86 \\
    \texttt{fr-en} &        39.73 &        85.82  & &        39.52 &        85.89  & &        44.31 &        87.86  & &        45.90 &        88.44 \\
    \texttt{he-en} &        33.60 &        82.69  & &        32.75 &        82.49  & &        37.90 &        84.99  & &        40.80 &        86.28 \\
    \texttt{hi-en} &        32.05 &        85.25  & &        30.64 &        84.80  & &        34.59 &        86.23  & &        37.23 &        87.49 \\
    \texttt{hr-en} &        33.19 &        84.60  & &        30.98 &        82.72  & &        37.43 &        86.69  & &        38.39 &        87.08 \\
    \texttt{hu-en} &        30.78 &        84.41  & &        29.70 &        83.71  & &        35.35 &        86.88  & &        36.91 &        87.50 \\
    \texttt{id-en} &        36.80 &        85.80  & &        35.30 &        84.92  & &        41.02 &        87.56  & &        43.13 &        88.20 \\
    \texttt{ja-en} &        20.33 &        82.40  & &        20.27 &        82.63  & &        25.18 &        85.46  & &        26.25 &        86.11 \\
    \texttt{ko-en} &        21.94 &        81.91  & &        22.28 &        82.26  & &        26.62 &        85.25  & &        28.08 &        85.78 \\
    \texttt{lt-en} &        28.39 &        82.46  & &        27.35 &        81.34  & &        33.36 &        85.19  & &        34.82 &        85.84 \\
    \texttt{nl-en} &        29.63 &        83.69  & &        28.37 &        82.53  & &        32.76 &        85.82  & &        33.83 &        86.28 \\
    \texttt{oc-en} &        43.82 &        78.81  & &        37.74 &        73.82  & &        46.37 &        79.64  & &        47.89 &        80.81 \\
    \texttt{pl-en} &        26.21 &        81.82  & &        24.84 &        80.57  & &        30.38 &        84.39  & &        31.49 &        84.94 \\
    \texttt{ps-en} &        15.72 &        68.46  & &        12.60 &        64.92  & &        17.95 &        71.62  & &        21.22 &        75.34 \\
    \texttt{pt-en} &        43.45 &        86.43  & &        42.62 &        85.94  & &        48.49 &        88.20  & &        50.15 &        88.69 \\
    \texttt{ru-en} &        29.54 &        82.72  & &        28.39 &        81.91  & &        34.30 &        84.87  & &        35.70 &        85.58 \\
    \texttt{sv-en} &        42.62 &        86.78  & &        42.07 &        86.01  & &        47.42 &        88.86  & &        49.34 &        89.41 \\
    \texttt{sw-en} &        33.36 &        79.05  & &        27.74 &        73.22  & &        35.24 &        80.18  & &        38.65 &        82.17 \\
    \texttt{ta-en} &        16.57 &        71.13  & &         8.13 &        59.07  & &         9.83 &        60.60  & &        13.82 &        66.54 \\
    \texttt{tr-en} &        32.04 &        85.21  & &        31.44 &        85.02  & &        35.69 &        87.03  & &        37.70 &        87.82 \\
    \texttt{vi-en} &        30.65 &        83.07  & &        30.23 &        83.00  & &        35.15 &        85.69  & &        36.65 &        86.45 \\
    \texttt{zh-en} &        22.02 &        81.58  & &        21.71 &        81.83  & &        27.09 &        84.76  & &        27.25 &        84.93 \\
\bottomrule
\end{tabular}
\caption{Translation quality according to spBLEU and COMET-22 of all the models and language pairs used in the English-centric \textsc{Flores} setup.}
\label{app:tab:quality_flores_setup_engcentric}
\end{table}

 \pagebreak

\begin{figure}[H]
    \centering
    \includegraphics[width=\textwidth]{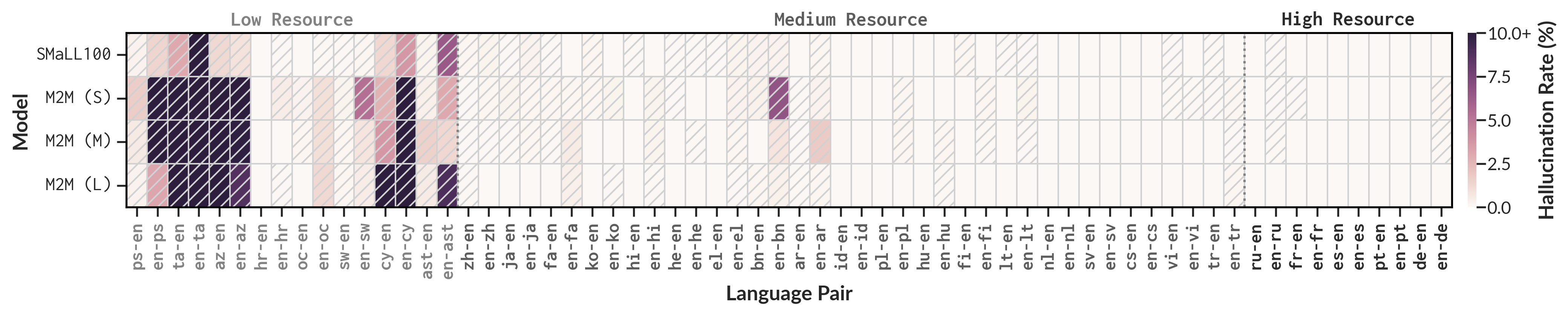}
    \caption{Heatmap of language-pair specific hallucination rates for each model in each of the translation directions considered in the English-centric \textsc{Flores} setup. Pattern-filled cells indicate at least one hallucination for a given model-LP combination.}
    \label{fig:lang_rates_engcentric}
\end{figure}

\paragraph{Hallucination rates for all language pairs.} We present a heatmap with hallucination rates for each model for all translation directions in Figure~\ref{fig:lang_rates_engcentric}.

\subsubsection{Evaluation with WMT.} 
\paragraph{Evaluation scores.} We present spBLEU and COMET-22 corpus-level scores in for the WMT setup in Table~\ref{app:quality_wmt_setup_engcentric}.

\begin{table}[H]
\renewcommand\arraystretch{1}
\scriptsize
\centering
\begin{tabular}{lrrcrrcrrcrr}
\toprule
\multirow{2}{*}{LP} &  \multicolumn{2}{c}{\texttt{SMaLL100}} & &  \multicolumn{2}{c}{\texttt{M2M (S)}} & & \multicolumn{2}{c}{\texttt{M2M (M)}} & & \multicolumn{2}{c}{\texttt{M2M (L)}}\\\cmidrule{2-3}\cmidrule{5-6}\cmidrule{8-9}\cmidrule{11-12}
& spBLEU &  COMET-22 & &  spBLEU &  COMET-22 & &  spBLEU &  COMET-22 & &  spBLEU &  COMET-22 \\
\midrule
\multicolumn{12}{c}{\texttt{en-xx} \textit{directions}}\\ \cdashlinelr{1-12}
\texttt{en-cs} &        23.90 &        81.68  & &        24.03 &        82.05  & &        28.61 &        86.85  & &        30.28 &        87.72 \\
    \texttt{en-de} &        35.92 &        77.29  & &        33.33 &        76.26  & &        42.85 &        83.10  & &        44.45 &        84.19 \\
    \texttt{en-et} &        25.22 &        82.80  & &        24.21 &        82.37  & &        30.50 &        87.66  & &        31.58 &        88.00 \\
    \texttt{en-fi} &        25.52 &        83.83  & &        25.55 &        84.45  & &        32.41 &        89.28  & &        34.60 &        90.02 \\
    \texttt{en-fr} &        37.37 &        81.54  & &        36.54 &        81.52  & &        41.61 &        85.23  & &        43.59 &        85.96 \\
    \texttt{en-ha} &         8.44 &        59.30  & &         3.02 &        39.47  & &         6.94 &        55.06  & &         9.86 &        61.35 \\
    \texttt{en-is} &        19.00 &        69.35  & &        14.36 &        62.44  & &        22.63 &        76.00  & &        25.60 &        77.91 \\
    \texttt{en-ja} &        16.25 &        81.69  & &        16.95 &        81.91  & &        19.11 &        83.86  & &        19.94 &        84.86 \\
    \texttt{en-lt} &        21.83 &        77.36  & &        20.93 &        77.76  & &        25.75 &        83.29  & &        26.51 &        83.72 \\
    \texttt{en-lv} &        17.16 &        73.95  & &        20.98 &        79.83  & &        26.49 &        85.10  & &        28.12 &        85.39 \\
    \texttt{en-ru} &        27.01 &        78.71  & &        26.91 &        78.76  & &        32.70 &        84.24  & &        34.79 &        85.50 \\
    \texttt{en-tr} &        24.15 &        81.64  & &        25.00 &        82.86  & &        26.14 &        84.89  & &        28.67 &        86.15 \\
    \texttt{en-uk} &        24.67 &        79.69  & &        24.53 &        78.99  & &        29.02 &        83.11  & &        29.27 &        83.65 \\
    \texttt{en-zh} &        15.50 &        74.39  & &        16.57 &        74.63  & &        19.53 &        79.04  & &        19.40 &        79.34 \\\midrule
    \multicolumn{12}{c}{\texttt{xx-en} \textit{directions}}\\ \cdashlinelr{1-12}
    \texttt{cs-en} &        28.08 &        80.64  & &        28.19 &        80.45  & &        31.84 &        82.73  & &        32.78 &        83.38 \\
    \texttt{de-en} &        34.35 &        77.89  & &        35.78 &        77.96  & &        39.86 &        80.62  & &        40.91 &        82.10 \\
    \texttt{et-en} &        27.92 &        82.14  & &        27.74 &        81.48  & &        32.79 &        84.98  & &        33.48 &        85.43 \\
    \texttt{fi-en} &        28.33 &        83.39  & &        28.08 &        82.54  & &        33.28 &        86.41  & &        33.76 &        86.78 \\
    \texttt{fr-en} &        34.11 &        83.17  & &        34.38 &        83.30  & &        38.56 &        85.52  & &        40.02 &        86.00 \\
    \texttt{ha-en} &        11.25 &        57.35  & &         7.25 &        48.77  & &        12.22 &        60.96  & &        13.77 &        63.71 \\
    \texttt{is-en} &        26.09 &        75.40  & &        23.04 &        70.47  & &        30.44 &        78.82  & &        31.62 &        80.83 \\
    \texttt{ja-en} &        12.54 &        71.53  & &        11.98 &        70.84  & &        14.21 &        73.19  & &        14.64 &        73.35 \\
    \texttt{lt-en} &        29.28 &        81.30  & &        29.52 &        80.96  & &        33.74 &        84.42  & &        34.16 &        84.46 \\
    \texttt{lv-en} &        20.14 &        77.19  & &        21.34 &        77.78  & &        24.62 &        81.58  & &        25.12 &        81.94 \\
    \texttt{ru-en} &        33.92 &        79.93  & &        35.15 &        75.36  & &        39.67 &        82.53  & &        40.59 &        82.85 \\
    \texttt{tr-en} &        24.69 &        80.50  & &        25.47 &        80.65  & &        27.54 &        82.41  & &        29.12 &        82.97 \\
    \texttt{uk-en} &        34.09 &        79.40  & &        32.19 &        77.18  & &        38.95 &        81.08  & &        38.82 &        81.12 \\
    \texttt{zh-en} &        19.60 &        75.02  & &        22.38 &        76.39  & &        27.95 &        79.42  & &        26.63 &        79.42 \\
\bottomrule
\end{tabular}
\caption{Translation quality according to spBLEU and COMET-22 of all the models and language pairs in our analysis of WMT benchmarks.}
\label{app:quality_wmt_setup_engcentric}
\end{table}

\begin{figure}[H]
    \centering
    \includegraphics[width=\textwidth]{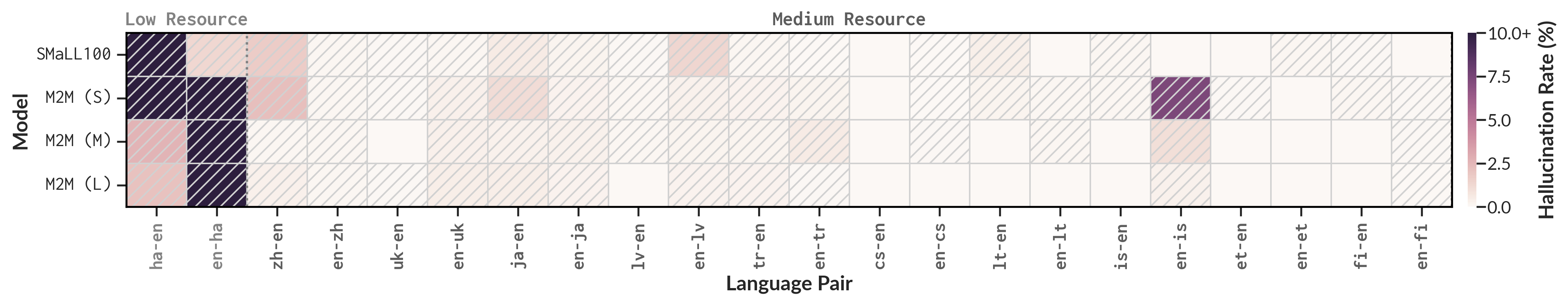}
    \caption{Heatmap of language-pair specific hallucination rates for each model in each of the translation directions considered in the WMT English-centric setup. Pattern-filled cells indicate at least one hallucination for a given model-LP combination.}
    \label{fig:lang_rates_wmtengcentric}
\end{figure}

\paragraph{Hallucination rates for all language pairs.} We present aggregated results in Table~\ref{tab:wmt_eng_centric} and a heatmap with hallucination rates for each model for all translation directions in Figure~\ref{fig:lang_rates_wmtengcentric}.\footnote{We do not present results for the high-resource language pairs as these were used as validation data to set the model-based ALTI+ thresholds.} Importantly, the trends regarding the prevalence of hallucinations analyzed in-depth in Section~\ref{sec:natural_halls} hold for the WMT language pairs as well.

\begin{table}[H]
\centering
\renewcommand\arraystretch{0.9}
\footnotesize
\begin{tabular}{>{\arraybackslash}m{1.45cm} >{\arraybackslash}m{1.75cm} >{\arraybackslash}m{1.15cm} c >{\raggedright\arraybackslash}m{1.75cm} >{\arraybackslash}m{1.15cm}}
\toprule
\multirow{2}{*}{\textsc{Model}} & \multicolumn{2}{c}{\textbf{\textsc{Low Resource}}} & & \multicolumn{2}{c}{\textbf{\textsc{Mid Resource}}}\\\cmidrule{2-3}\cmidrule{5-6}
& LP Fraction & Rate (\%) & & \multicolumn{1}{l}{{LP Fraction}} & Rate (\%)\\\midrule
\texttt{SMaLL100} & \begin{tikzpicture}[font=\scriptsize] 
 \draw [fill=tikz_red] (0,0) rectangle (2*6/2*0.15, 0.15); 
 \draw [fill=tikz_gray] (2*6/2*0.15, 0) rectangle (6*0.15, 0.15); 
 \node [left, color=tikz_red] at (0, 0.1) {\textbf{\scriptsize{2}}\textcolor{black!50}{/2}}; 
 \end{tikzpicture} & $5.658$ & &\begin{tikzpicture}[font=\scriptsize] 
 \draw [fill=tikz_red] (0,0) rectangle (15*6/20*0.15, 0.15); 
 \draw [fill=tikz_gray] (15*6/20*0.15, 0) rectangle (6*0.15, 0.15); 
 \node [left, color=tikz_red] at (0, 0.1) {\textbf{\scriptsize{15}}\textcolor{black!50}{/20}}; 
 \end{tikzpicture} & $0.228_{\textcolor{black!50}{\tiny{0.07}}}$ \\ \texttt{M2M (S)} & \begin{tikzpicture}[font=\scriptsize] 
 \draw [fill=tikz_red] (0,0) rectangle (2*6/2*0.15, 0.15); 
 \draw [fill=tikz_gray] (2*6/2*0.15, 0) rectangle (6*0.15, 0.15); 
 \node [left, color=tikz_red] at (0, 0.1) {\textbf{\scriptsize{2}}\textcolor{black!50}{/2}}; 
 \end{tikzpicture} & $47.57$ & &\begin{tikzpicture}[font=\scriptsize] 
 \draw [fill=tikz_red] (0,0) rectangle (18*6/20*0.15, 0.15); 
 \draw [fill=tikz_gray] (18*6/20*0.15, 0) rectangle (6*0.15, 0.15); 
 \node [left, color=tikz_red] at (0, 0.1) {\textbf{\scriptsize{18}}\textcolor{black!50}{/20}}; 
 \end{tikzpicture} & $0.454_{\textcolor{black!50}{\tiny{0.15}}}$ \\\texttt{M2M (M)} & \begin{tikzpicture}[font=\scriptsize] 
 \draw [fill=tikz_red] (0,0) rectangle (2*6/2*0.15, 0.15); 
 \draw [fill=tikz_gray] (2*6/2*0.15, 0) rectangle (6*0.15, 0.15); 
 \node [left, color=tikz_red] at (0, 0.1) {\textbf{\scriptsize{2}}\textcolor{black!50}{/2}}; 
 \end{tikzpicture} & $27.69$ & &\begin{tikzpicture}[font=\scriptsize] 
 \draw [fill=tikz_red] (0,0) rectangle (13*6/20*0.15, 0.15); 
 \draw [fill=tikz_gray] (13*6/20*0.15, 0) rectangle (6*0.15, 0.15); 
 \node [left, color=tikz_red] at (0, 0.1) {\textbf{\scriptsize{13}}\textcolor{black!50}{/20}}; 
 \end{tikzpicture} & $0.157_{\textcolor{black!50}{\tiny{0.10}}}$ \\\texttt{M2M (L)} & \begin{tikzpicture}[font=\scriptsize] 
 \draw [fill=tikz_red] (0,0) rectangle (2*6/2*0.15, 0.15); 
 \draw [fill=tikz_gray] (2*6/2*0.15, 0) rectangle (6*0.15, 0.15); 
 \node [left, color=tikz_red] at (0, 0.1) {\textbf{\scriptsize{2}}\textcolor{black!50}{/2}}; 
 \end{tikzpicture} & $9.364$ & &\begin{tikzpicture}[font=\scriptsize] 
 \draw [fill=tikz_red] (0,0) rectangle (11*6/20*0.15, 0.15); 
 \draw [fill=tikz_gray] (11*6/20*0.15, 0) rectangle (6*0.15, 0.15); 
 \node [left, color=tikz_red] at (0, 0.1) {\textbf{\scriptsize{11}}\textcolor{black!50}{/20}}; 
 \end{tikzpicture} & $0.124_{\textcolor{black!50}{\tiny{0.05}}}$ \\\bottomrule
\end{tabular}
\caption{Fraction of LPs on the WMT English-centric setup for which models produce at least one hallucination, and average hallucination rate (and median, in subscript) across all LPs at each resource level. We do not present the median for the low-resource LPs as we only evaluate on 2 translation directions.}
\label{tab:wmt_eng_centric}
\end{table}

\begin{figure}[H]
    \centering
    \includegraphics[width=\textwidth]{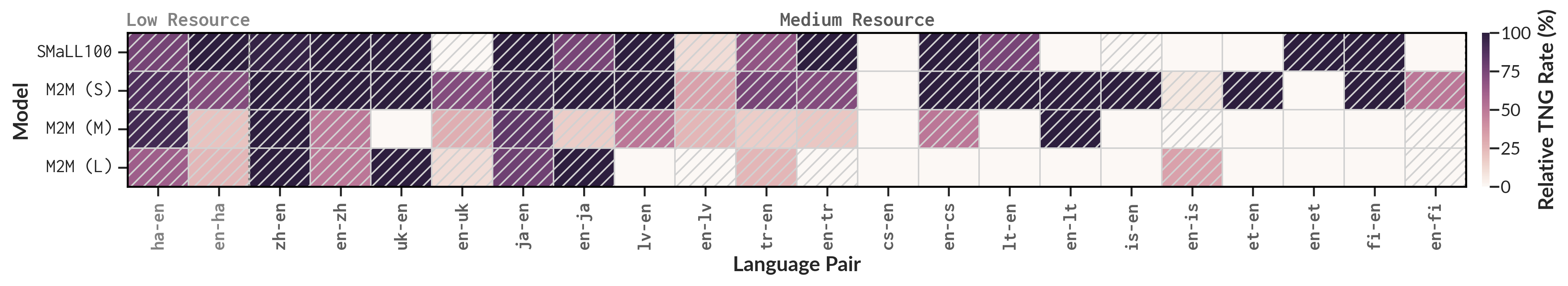}
    \caption{ Heatmap of the percentage of hallucinations detected with TNG (oscillatory hallucinations) among all detected hallucinations in the WMT setup. Pattern-filled cells indicate at least one hallucination detected for a given model-LP combination.}
    \label{fig:tng_hall_rates_wmt}
\end{figure}

\paragraph{Prevalence of oscillatory hallucinations.} We present a heatmap with the relative prevalence of hallucinations detected only by TNG~(oscillatory hallucinations) in Figure~\ref{fig:tng_hall_rates_wmt}.

\subsection{Beyond English-Centric Translation}
\label{app:subsec:nonengcentric}
\paragraph{Evaluation scores.} We present spBLEU and COMET-22 corpus-level scores in Table~\ref{app:tab:quality_flores_setup_nonengcentric}.

\begin{figure}[H]
    \centering
    \includegraphics[width=\textwidth]{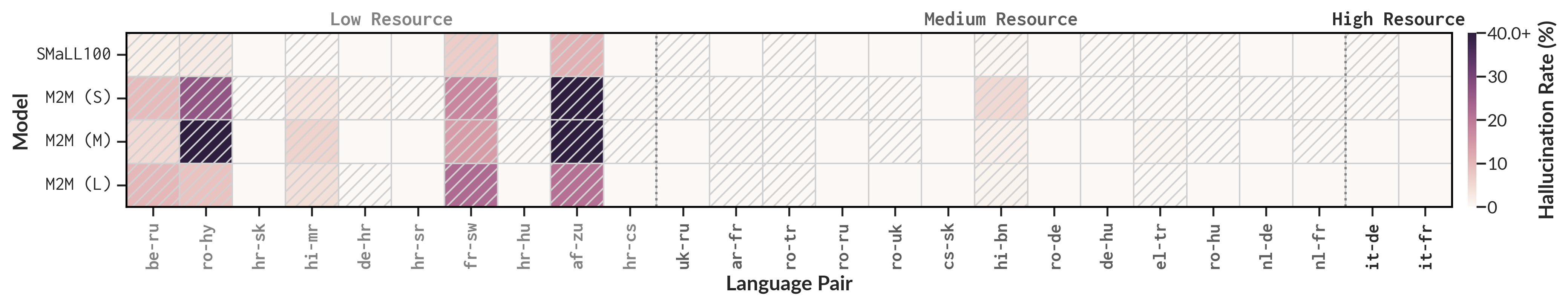}
    \caption{Heatmap of language-pair specific hallucination rates for each model in each of the translation directions considered in the non-English-centric setup. Pattern-filled cells indicate at least one hallucination for a given model-LP combination.}
    \label{fig:lang_rates_nonengcentric}
\end{figure}

\paragraph{Hallucination rates for all language pairs.} We present a heatmap with hallucination rates for each model for all translation directions in Figure~\ref{fig:lang_rates_nonengcentric}.

\begin{figure}[H]
    \centering
    \includegraphics[width=\textwidth]{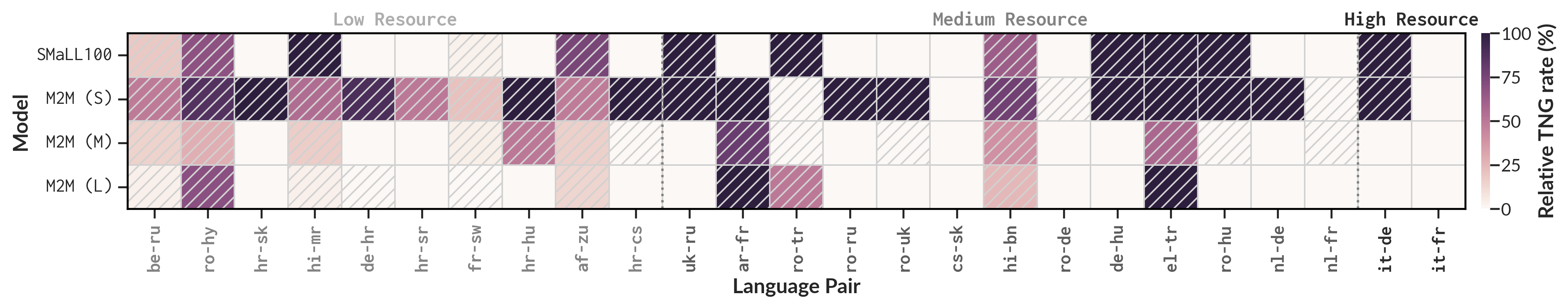}
    \caption{ Heatmap of the percentage of hallucinations detected with TNG (oscillatory hallucinations) among all detected hallucinations in the non-English-centric setup. Pattern-filled cells indicate at least one hallucination detected for a given model-LP combination.}
    \label{fig:tng_hall_rates_nonengcentric}
\end{figure}

\paragraph{Prevalence of oscillatory hallucinations.} We present a heatmap with the relative prevalence of hallucinations detected only by TNG~(oscillatory hallucinations) in Figure~\ref{fig:tng_hall_rates_nonengcentric}. The trends follow those presented in the analysis in the main text~(see Section~\ref{subsec:analysis_eng_centric_flores}).

\begin{table}[H]
\renewcommand\arraystretch{1}
\scriptsize
\centering
\begin{tabular}{lrrcrrcrrcrr}
\toprule
\multirow{2}{*}{LP} &  \multicolumn{2}{c}{\texttt{SMaLL100}} & &  \multicolumn{2}{c}{\texttt{M2M (S)}} & & \multicolumn{2}{c}{\texttt{M2M (M)}} & & \multicolumn{2}{c}{\texttt{M2M (L)}}\\\cmidrule{2-3}\cmidrule{5-6}\cmidrule{8-9}\cmidrule{11-12}
& spBLEU &  COMET-22 & &  spBLEU &  COMET-22 & &  spBLEU &  COMET-22 & &  spBLEU &  COMET-22 \\
\midrule
\texttt{af-zu} &         6.89 &        55.73  & &         4.09 &        43.06  & &         8.30 &        59.19  & &        13.83 &        66.00 \\
    \texttt{ar-fr} &        28.49 &        77.18  & &        27.80 &        76.86  & &        28.45 &        76.60  & &        34.99 &        82.07 \\
    \texttt{be-ru} &        17.40 &        82.46  & &        15.60 &        78.75  & &        19.31 &        85.07  & &        18.34 &        83.72 \\
    \texttt{cs-sk} &        32.63 &        90.55  & &        33.34 &        91.05  & &        35.90 &        92.19  & &        36.15 &        92.37 \\
    \texttt{de-hr} &        24.69 &        84.98  & &        23.37 &        84.10  & &        28.81 &        88.90  & &        29.52 &        89.18 \\
    \texttt{de-hu} &        23.49 &        80.98  & &        23.26 &        81.30  & &        28.67 &        86.63  & &        30.01 &        87.09 \\
    \texttt{el-tr} &        19.09 &        79.00  & &        19.83 &        79.76  & &        21.66 &        82.35  & &        23.47 &        83.58 \\
    \texttt{fr-sw} &        23.23 &        75.61  & &        15.62 &        68.06  & &        21.52 &        74.22  & &        22.72 &        73.41 \\
    \texttt{hi-bn} &        24.12 &        79.42  & &        16.19 &        71.90  & &        23.63 &        79.58  & &        25.63 &        81.06 \\
    \texttt{hi-mr} &        14.79 &        65.69  & &        10.59 &        61.41  & &        13.68 &        63.22  & &        18.13 &        68.16 \\
    \texttt{hr-cs} &        26.07 &        86.59  & &        25.87 &        85.82  & &        30.79 &        90.00  & &        31.11 &        90.32 \\
    \texttt{hr-hu} &        22.22 &        81.22  & &        21.81 &        80.44  & &        27.24 &        86.25  & &        27.93 &        86.44 \\
    \texttt{hr-sk} &        27.72 &        86.54  & &        26.63 &        85.40  & &        31.84 &        89.75  & &        31.89 &        90.03 \\
    \texttt{hr-sr} &        27.95 &        87.37  & &        28.34 &        86.75  & &        30.76 &        89.38  & &        26.30 &        88.05 \\
    \texttt{it-de} &        23.74 &        78.69  & &        21.89 &        76.74  & &        27.91 &        83.36  & &        29.66 &        84.30 \\
    \texttt{it-fr} &        32.63 &        82.19  & &        32.32 &        81.94  & &        36.74 &        85.39  & &        37.47 &        85.93 \\
    \texttt{nl-de} &        22.64 &        80.23  & &        21.65 &        78.61  & &        26.82 &        84.08  & &        27.97 &        84.84 \\
    \texttt{nl-fr} &        27.50 &        77.88  & &        26.73 &        77.09  & &        31.41 &        82.03  & &        33.08 &        82.86 \\
    \texttt{ro-de} &        27.89 &        80.45  & &        25.74 &        78.81  & &        33.18 &        84.76  & &        34.73 &        85.53 \\
    \texttt{ro-hu} &        23.75 &        80.91  & &        23.47 &        80.67  & &        28.95 &        86.30  & &        29.79 &        86.66 \\
    \texttt{ro-hy} &        11.22 &        63.12  & &         5.06 &        51.47  & &         6.24 &        51.55  & &         9.47 &        61.76 \\
    \texttt{ro-ru} &        24.61 &        83.48  & &        23.95 &        82.94  & &        29.33 &        87.24  & &        30.99 &        88.06 \\
    \texttt{ro-tr} &        23.46 &        81.17  & &        24.30 &        81.79  & &        26.16 &        84.49  & &        27.77 &        85.48 \\
    \texttt{ro-uk} &        24.77 &        83.65  & &        23.93 &        82.96  & &        28.99 &        87.36  & &        29.23 &        87.67 \\
    \texttt{uk-ru} &        30.01 &        89.44  & &        29.96 &        89.10  & &        33.28 &        91.65  & &        33.56 &        91.56 \\
\bottomrule
\end{tabular}
\caption{Translation quality according to spBLEU and COMET-22 of all the models and language pairs used in the non-English-centric setup.}
\label{app:tab:quality_flores_setup_nonengcentric}
\end{table}

\newpage
\subsection{Specialized Domains}
\label{app:subsec:specialized_domain}
\paragraph{Evaluation scores.} We present spBLEU and COMET-22 corpus-level scores in Table~\ref{app:tab:quality_flores_setup_TICO}.

\begin{table}[H]
\renewcommand\arraystretch{0.95}
\scriptsize
\centering
\begin{tabular}{lrrcrrcrrcrr}
\toprule
\multirow{2}{*}{LP} &  \multicolumn{2}{c}{\texttt{SMaLL100}} & &  \multicolumn{2}{c}{\texttt{M2M (S)}} & & \multicolumn{2}{c}{\texttt{M2M (M)}} & & \multicolumn{2}{c}{\texttt{M2M (L)}}\\\cmidrule{2-3}\cmidrule{5-6}\cmidrule{8-9}\cmidrule{11-12}
& spBLEU &  COMET-22 & &  spBLEU &  COMET-22 & &  spBLEU &  COMET-22 & &  spBLEU &  COMET-22 \\
\midrule
\multicolumn{12}{c}{\texttt{en-xx} \textit{directions}}\\ \cdashlinelr{1-12}
\texttt{en-ar} &        26.00 &        79.41  & &        25.21 &        79.50  & &        16.62 &        74.15  & &        29.59 &        82.84 \\
    \texttt{en-fr} &        37.47 &        77.60  & &        37.39 &        77.23  & &        42.81 &        80.11  & &        44.48 &        80.68 \\
    \texttt{en-hi} &        40.49 &        75.45  & &        38.87 &        75.35  & &        39.97 &        76.47  & &        43.02 &        77.99 \\
    \texttt{en-id} &        47.64 &        89.00  & &        44.01 &        88.15  & &        49.95 &        90.04  & &        52.33 &        90.55 \\
    \texttt{en-ps} &        12.02 &        59.99  & &        10.74 &        56.89  & &        16.92 &        65.32  & &        22.60 &        71.37 \\
    \texttt{en-pt} &        48.89 &        86.96  & &        48.61 &        86.79  & &        53.72 &        88.87  & &        55.26 &        89.22 \\
    \texttt{en-ru} &        30.43 &        82.34  & &        29.67 &        81.84  & &        36.48 &        86.07  & &        38.52 &        86.75 \\
    \texttt{en-sw} &        33.00 &        78.23  & &        23.98 &        71.91  & &        31.97 &        78.92  & &        35.39 &        80.06 \\
    \texttt{en-zh} &        19.77 &        80.22  & &        20.63 &        80.13  & &        24.74 &        84.18  & &        23.47 &        83.58 \\\midrule
    \multicolumn{12}{c}{\texttt{xx-en} \textit{directions}}\\ \cdashlinelr{1-12}
    \texttt{ar-en} &        31.02 &        81.63  & &        28.72 &        80.95  & &        29.38 &        80.26  & &        37.00 &        84.37 \\
    \texttt{fr-en} &        36.68 &        80.89  & &        36.67 &        80.69  & &        40.53 &        81.88  & &        42.09 &        82.24 \\
    \texttt{hi-en} &        42.42 &        86.46  & &        41.24 &        86.06  & &        45.62 &        87.21  & &        48.45 &        88.19 \\
    \texttt{id-en} &        43.59 &        87.24  & &        42.09 &        86.38  & &        48.34 &        88.49  & &        50.00 &        89.09 \\
    \texttt{ps-en} &        24.26 &        73.65  & &        19.70 &        69.89  & &        26.95 &        76.76  & &        31.80 &        80.40 \\
    \texttt{pt-en} &        49.06 &        87.48  & &        48.65 &        87.05  & &        53.74 &        88.73  & &        55.43 &        89.03 \\
    \texttt{ru-en} &        32.99 &        83.28  & &        31.05 &        82.74  & &        36.49 &        84.66  & &        37.96 &        85.13 \\
    \texttt{sw-en} &        34.40 &        78.70  & &        30.20 &        74.02  & &        36.67 &        79.61  & &        39.00 &        81.02 \\
    \texttt{zh-en} &        27.77 &        82.25  & &        27.88 &        82.19  & &        32.54 &        84.63  & &        32.86 &        84.82 \\
\bottomrule
\end{tabular}
\caption{Translation quality according to spBLEU and COMET-22 of all the models and language pairs used in the specialized domain data setup with the TICO dataset.}
\label{app:tab:quality_flores_setup_TICO}
\end{table}

\begin{figure}[H]
    \centering
    \includegraphics[width=\textwidth]{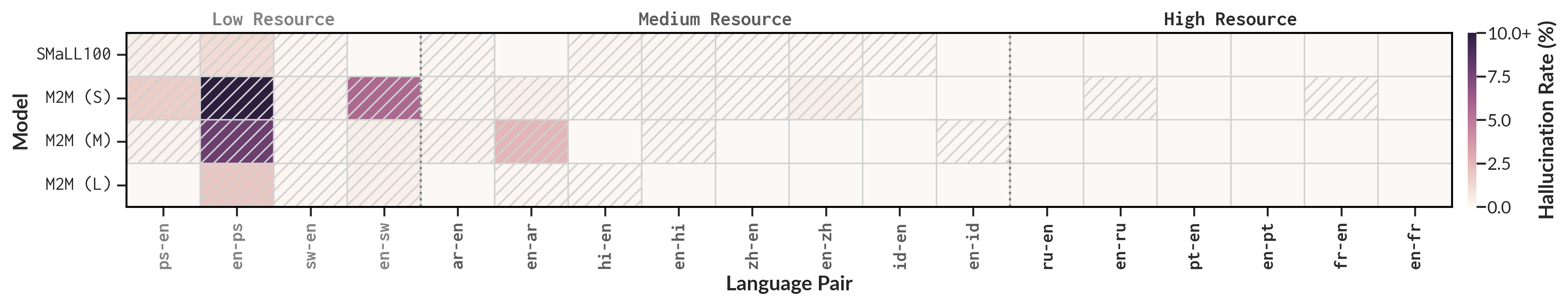}
    \caption{Heatmap of language-pair specific hallucination rates for each model in each of the translation directions considered in the specialized domain setup with TICO data. Pattern-filled cells indicate at least one hallucination for a given model-LP combination.}
    \label{fig:lang_rates_TICO}
\end{figure}

\paragraph{Hallucination rates for all language pairs.} We present a heatmap with hallucination rates for each model for all translation directions in Figure~\ref{fig:lang_rates_TICO}.

\begin{figure}[H]
    \centering
    \includegraphics[width=\textwidth]{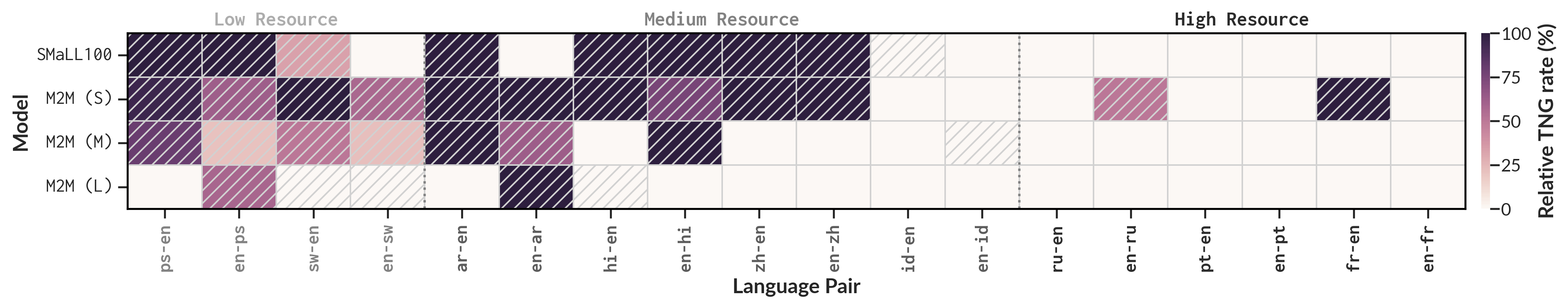}
    \caption{ Heatmap of the percentage of hallucinations detected with TNG (oscillatory hallucinations) among all hallucinations in the specialized domain setup with TICO data. Pattern-filled cells indicate at least one natural hallucination detected for a given model-LP combination.}
    \label{fig:tng_hall_rates_tico}
\end{figure}

\paragraph{Prevalence of oscillatory hallucinations.} We present a heatmap with the relative prevalence of hallucinations detected only by TNG~(oscillatory hallucinations) in Figure~\ref{fig:tng_hall_rates_tico}. Similarly to the non-English-centric setup, the trends follow those presented in the analysis in the main text~(see Section~\ref{subsec:analysis_eng_centric_flores}).

\section{Fallback System Analysis}
\label{app:fallbacksystemanalysis}
\begin{sidewaysfigure}
    \centering
    \includegraphics[width=\textwidth]{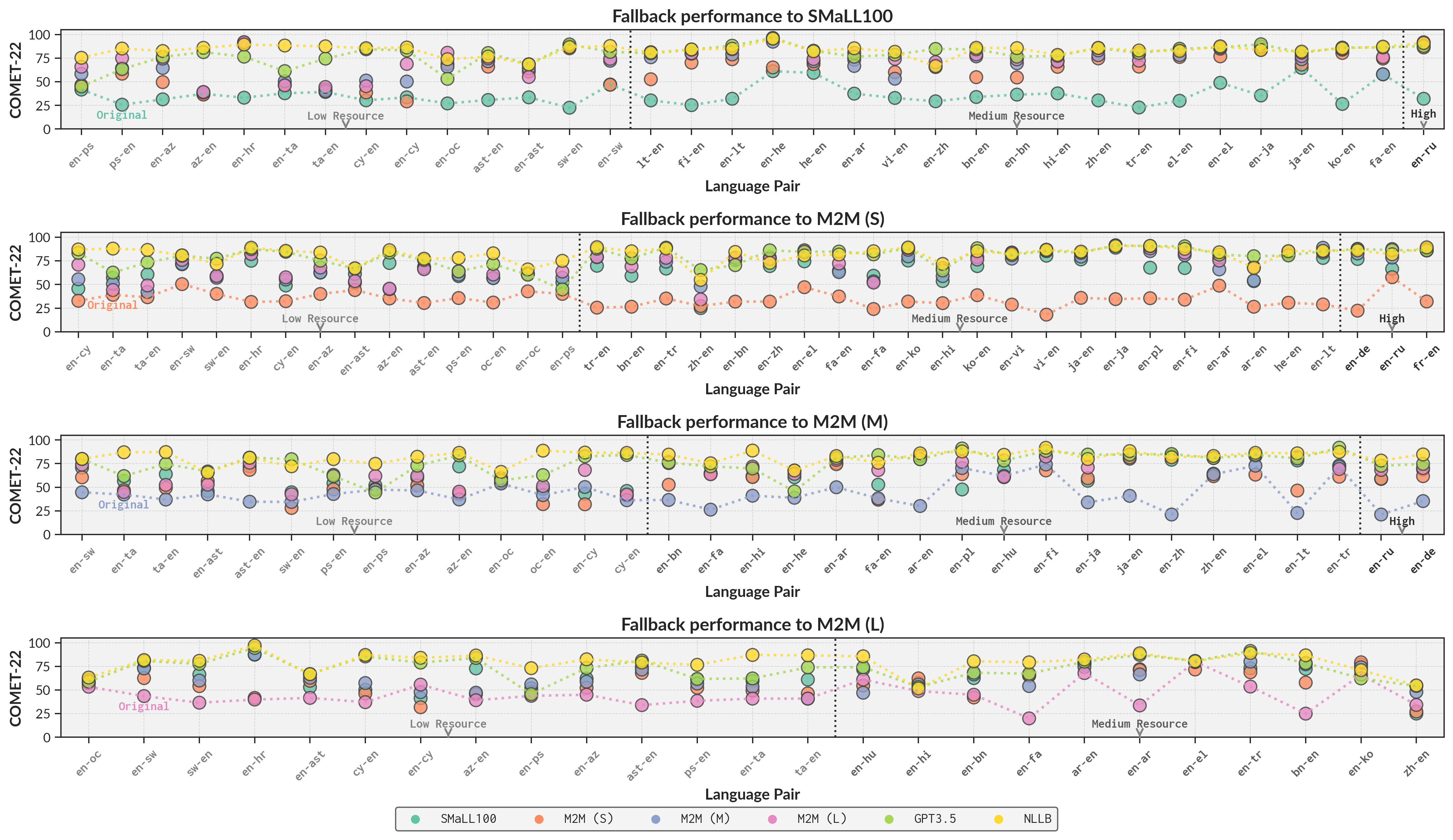}
    \caption{Overall translation quality analysis with COMET-22 scores on the original model hallucinated translations (represented with dashed lines) across different resource levels for each language pair in the \textsc{Flores} English-centric setup.}
    \label{fig:fallback_engcentric_lpspecific_comet22}
\end{sidewaysfigure}

\paragraph{English-Centric directions.} We present COMET-22 scores on the original model hallucinated translations for every translation direction in Figure~\ref{fig:fallback_engcentric_lpspecific_comet22}.

\begin{figure}[H]
\begin{subfigure}[b]{0.495\textwidth}
\centering
\includegraphics[scale=0.645]{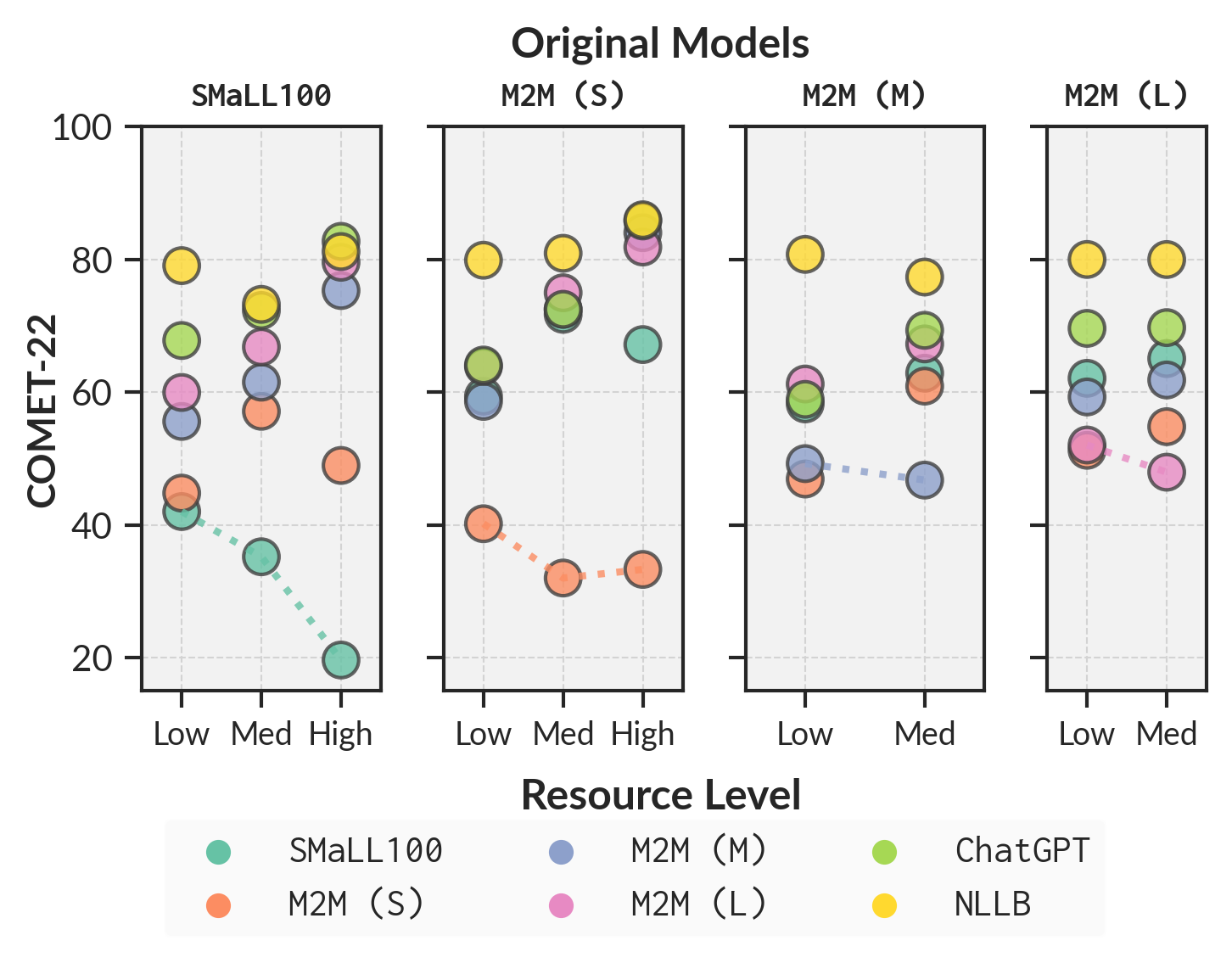}
\caption{Translation quality.}
\label{fig:flores_nonengcentric_fallback_quality}
\end{subfigure} \quad 
\begin{subfigure}[b]{0.495\textwidth}
\centering
\includegraphics[scale=0.555]{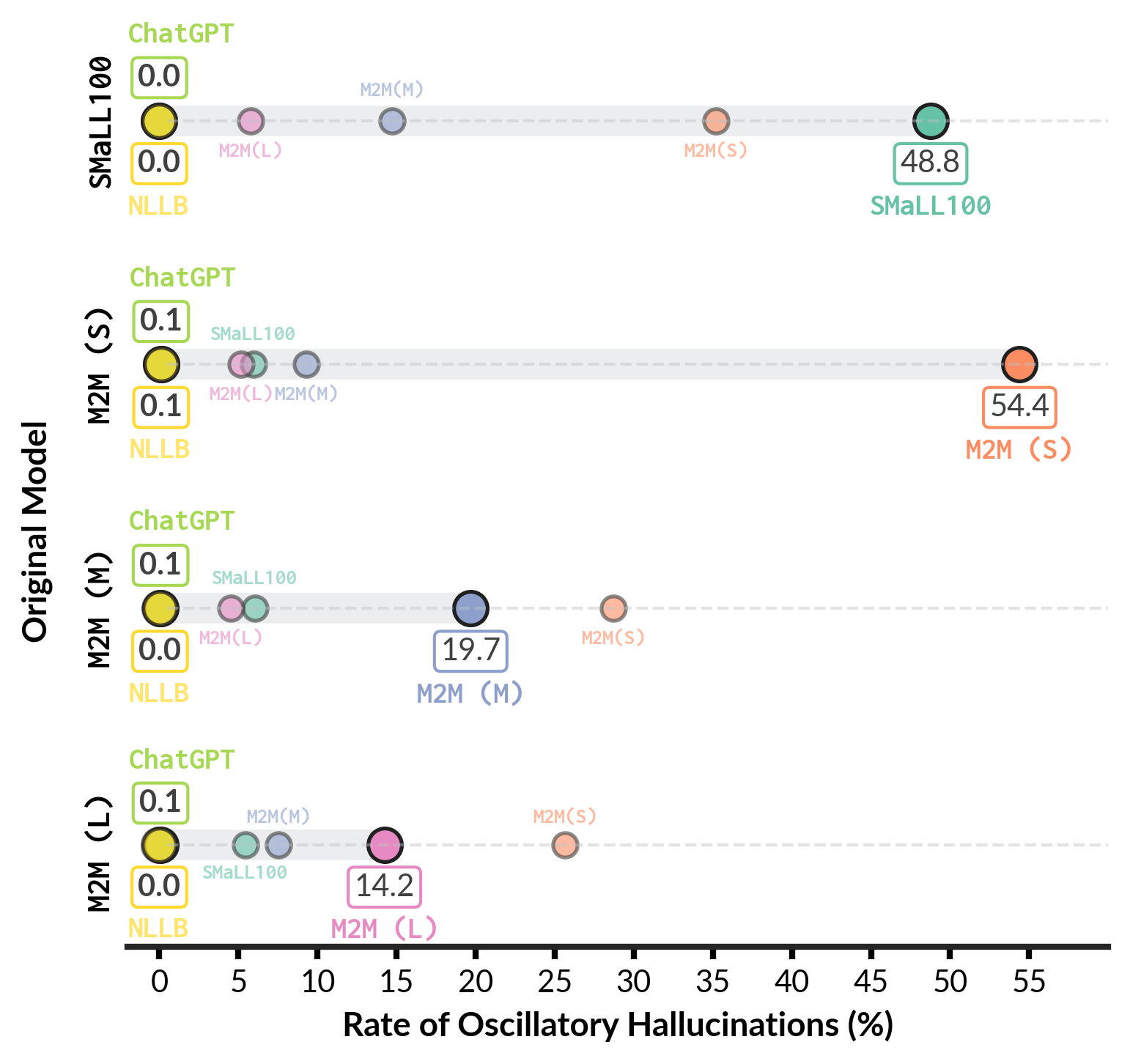}
\caption{Prevalence of oscillatory hallucinations.}
\label{fig:flores_nonengcentric_fallback_oscillatory}
\end{subfigure}
\caption{Fallback system analysis for the non-English-centric setup. We analyse overall translation quality improvements on the original model hallucinated translations~(represented with dashed lines) across different resource levels via COMET-22 scores in~(a), and overall prevalence of oscillatory hallucinations among the fallback translations in~(b).}
\label{fig:flores_nonengcentric_reversal_quality_osc}
\end{figure}

\begin{sidewaysfigure}
    \centering
    \includegraphics[width=\textwidth]{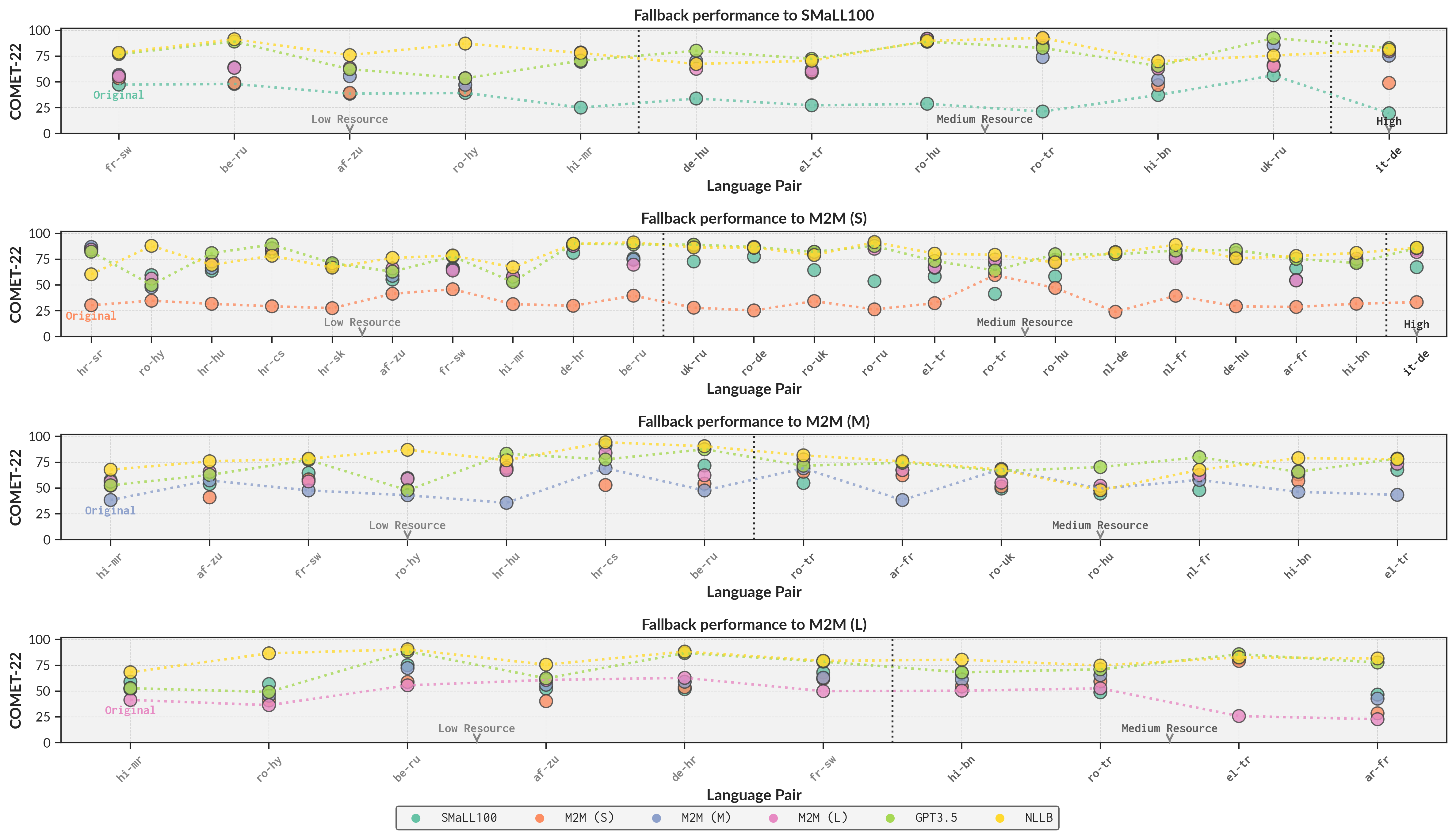}
    \caption{Overall translation quality analysis with COMET-22 scores on the original model hallucinated translations (represented with dashed lines) across different resource levels for each language pair in the non-English-centric setup.}
    \label{fig:fallback_nonengcentric_lpspecific_comet22}
\end{sidewaysfigure}

\paragraph{Non-English-Centric directions.} We present the results on employing external models as fallback systems in Figure~\ref{fig:flores_nonengcentric_reversal_quality_osc}, and COMET-22 scores on the original model hallucinated translations for each translation direction in the non-English-centric setup in Figure~\ref{fig:fallback_nonengcentric_lpspecific_comet22}. Importantly, the trends analyzed in-depth in the main text largely hold in this setup as well. In contrast with the analysis for the English-centric setup, \texttt{ChatGPT} struggles to outperform all other models in Non-English-Centric directions, especially in low and mid-resource levels. This is expected: as \texttt{ChatGPT} was trained on a non-parallel heavily English-centric corpus, it is likely that it struggles more with translation directions that do not include English.

\newpage
\section{Examples of Natural Hallucinations}
\begin{figure}[H]
    \includegraphics[width=\textwidth]{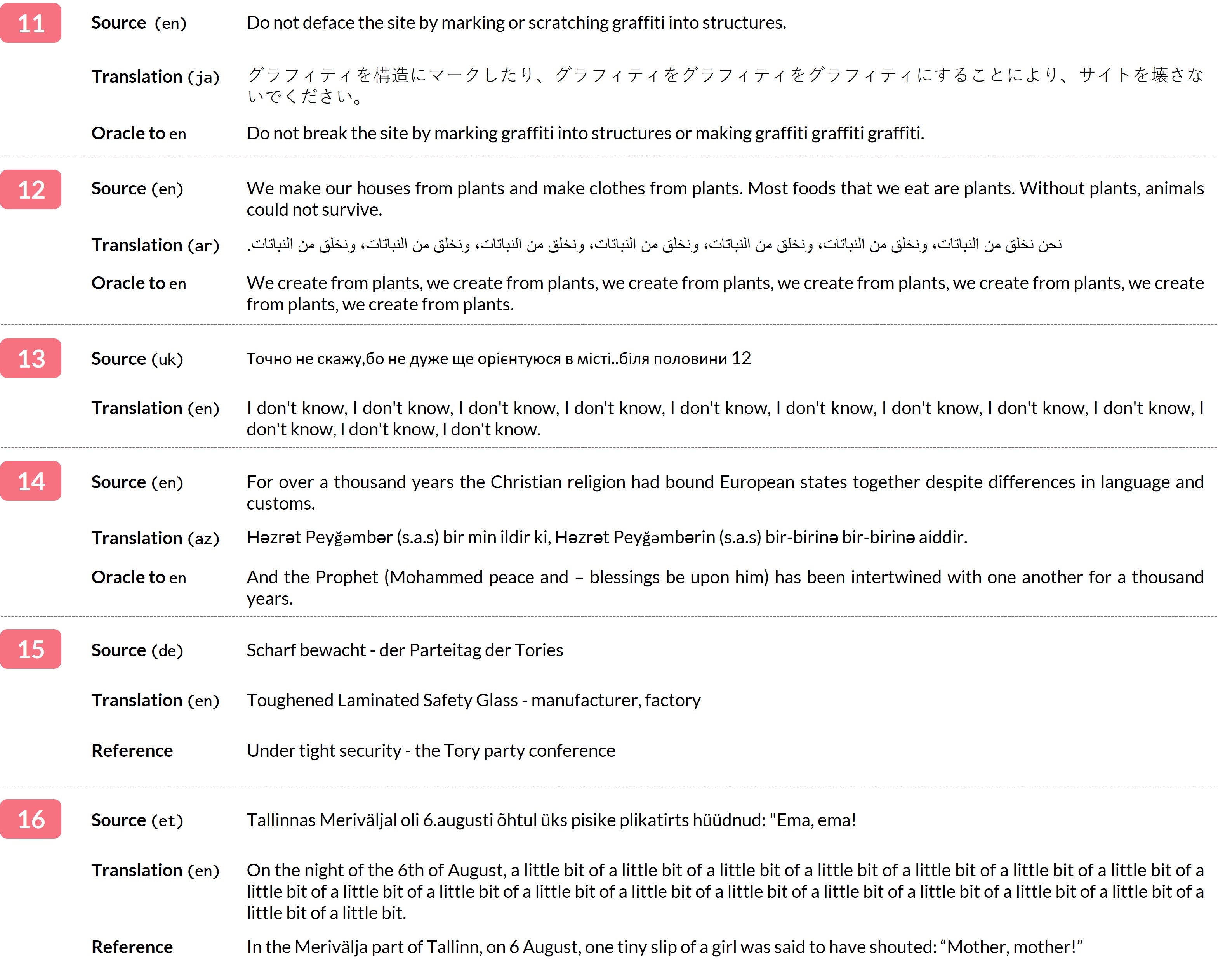}
    \caption{Examples of hallucinations generated by the M2M models in English-centric directions. We present an oracle English translation of the generated hallucination with Bing Microsoft Translator (\textsc{Oracle to \texttt{en}}) when necessary.}
    \label{fig:app:m2m_halls}
\end{figure}

We provide several examples of English-centric natural hallucinations generated by the M2M models in Figure~\ref{fig:app:m2m_halls}, which spans 2 pages. 

\begin{figure}[H]
    \centering
    \includegraphics[width=\textwidth]{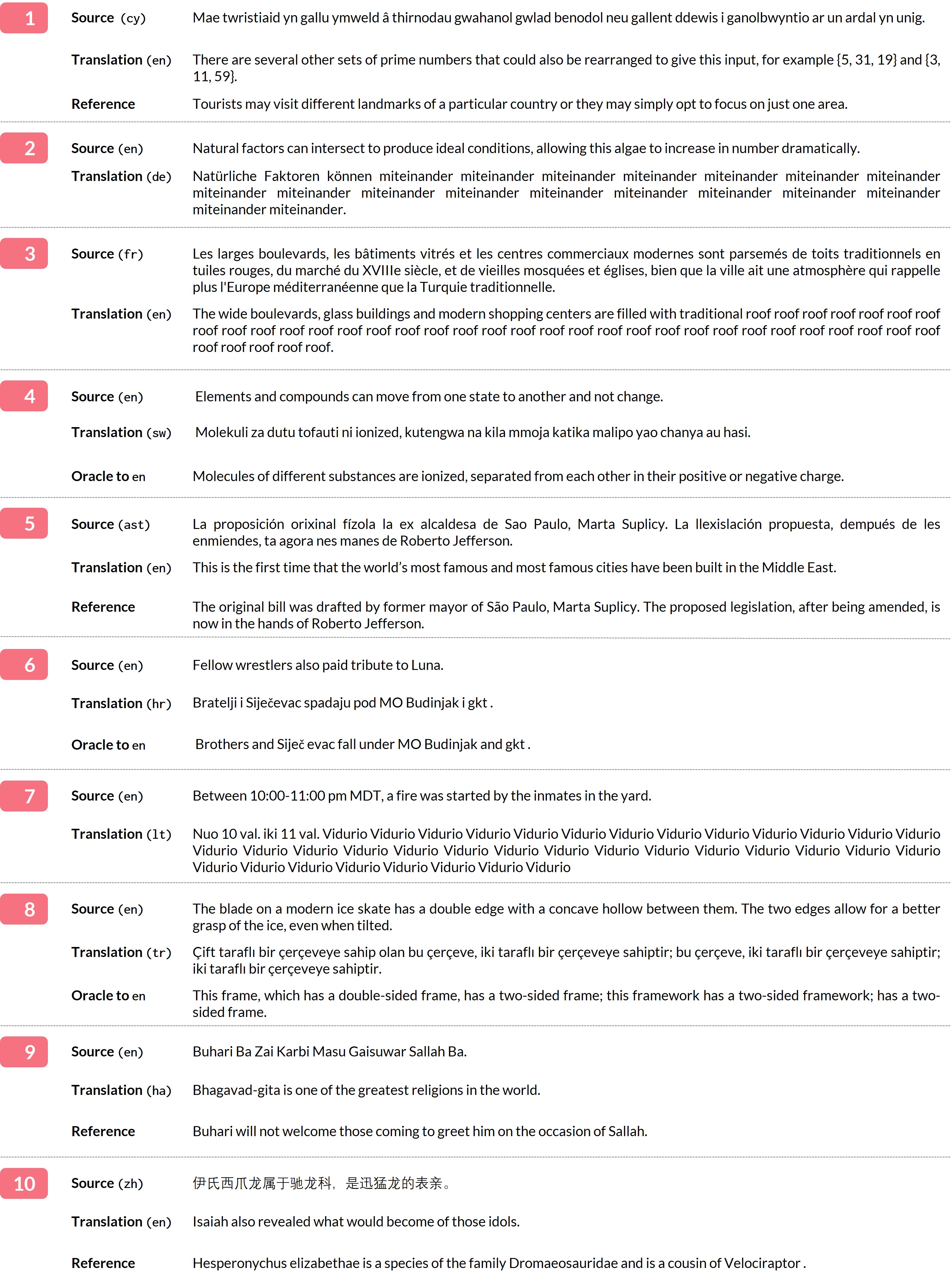}
\end{figure}\newpage

\end{document}